\RequirePackage{fix-cm}
\documentclass{article}

\usepackage{iclr2027_conference,times}
\usepackage{textcomp}
\usepackage{amsmath,amssymb,amsthm,bm}
\usepackage{dsfont}
\usepackage{booktabs}
\usepackage{array}
\usepackage{graphicx}
\usepackage{tikz}
\usepackage{xcolor}
\usepackage{colortbl}
\usepackage{hyperref}
\usepackage{microtype}
\usepackage{color}

\definecolor{rcSlateDark}{HTML}{1E293B}
\definecolor{rcSlateMid}{HTML}{475569}
\definecolor{rcBorder}{HTML}{CBD5E1}
\definecolor{rcHeaderBg}{HTML}{F1F5F9}
\definecolor{rcSubHdrBg}{HTML}{F8FAFC}
\definecolor{rcOursBg}{HTML}{EFF6FF}
\definecolor{rcOursHdr}{HTML}{DBEAFE}
\definecolor{rcGreen}{HTML}{0D7A5F}
\definecolor{rcGreenBg}{HTML}{D1FAE5}
\definecolor{rcBlue}{HTML}{1D4ED8}
\definecolor{rcBlueBg}{HTML}{DBEAFE}
\definecolor{rcPurple}{HTML}{6D28D9}
\definecolor{rcPurpleBg}{HTML}{EDE9FE}
\definecolor{rcRed}{HTML}{B42318}

\newcommand{\rcTwo}[2]{%
  \begin{tabular}[t]{@{}l@{}}#1\\[-0.5pt]#2\end{tabular}%
}
\newcommand{\rcRowSep}{%
  \arrayrulecolor{black!45}\specialrule{0.45pt}{3.5pt}{3.7pt}\arrayrulecolor{black}%
}
\newcommand{\rcBoldMu}{\pdfliteral direct {2 Tr 0.42 w 1 J 1 j}\textmu\pdfliteral direct {0 Tr}}

\newsavebox{\promptboxsave}
\newenvironment{promptbox}{%
  \par\medskip\noindent
  \setlength{\fboxsep}{8pt}%
  \begin{lrbox}{\promptboxsave}%
  \begin{minipage}{\dimexpr\linewidth-2\fboxsep-2\fboxrule\relax}%
  \footnotesize}%
  {\end{minipage}\end{lrbox}%
  \setlength{\fboxsep}{8pt}%
  \fcolorbox{black!55}{black!4}{\usebox{\promptboxsave}}%
  \par\medskip}

\newtheorem{assumption}{Assumption}

\newtheorem{proposition}{Proposition}

\newtheorem{corollary}{Corollary}

\title{RegimeCouncil: Paired-Evidence Routing for\\
Memory Conflict Governance in LLM Agents}

\author{Qingcan Kang$^{1*}$, Mingyang LIU$^{2*}$, Shixiong Kai$^{1}$, Tao Zhong$^{1}$, Mingxuan Yuan$^{1\dagger}$ \\
$^{1}$Huawei Noah's Ark Lab, Hong Kong \\
$^{2}$City University of Hong Kong, Hong Kong \\
\texttt{\{kang.qingcan, kaishixiong, zhongtao5, Yuan.Mingxuan\}@huawei.com} \\
\texttt{mingyaliu8-c@my.cityu.edu.hk}
}

\iclrfinalcopy

\begin{document}

\maketitle

\begingroup
\renewcommand\thefootnote{}%
\csname NoHyper\endcsname
\footnotetext{$^{*}$Equal contribution.\quad $^{\dagger}$Corresponding author.}%
\csname endNoHyper\endcsname
\addtocounter{footnote}{0}%
\endgroup

\begin{abstract}
A fundamental challenge in long-term LLM agent memory is not only storing and
retrieving facts, but deciding how to use them when they conflict with a
changing world. Facts often become stale without explicit deletion, so conflicts
appear before the store is refreshed. Delayed feedback may show that using an
outdated record hurt on one query, but not whether suppression should remain
fact-local (confined to that fact) or become store-wide (making the agent more
conservative elsewhere). Existing systems typically fix one sharing scope at
design time, whereas the appropriate scope is unknown at decision time and must
be inferred online from delayed outcomes. We formulate this online choice as
\emph{memory conflict governance} under hidden validity and delayed feedback, and
propose RegimeCouncil, a lightweight router that arbitrates between a
fact-local advisor and a store-wide advisor without retraining the LLM. Each
memory-assisted response is scored against a no-memory anchor via off-path
paired comparison or inverse-propensity estimation from executed actions alone:
a local advisor tracks the queried fact, a global advisor pools recent outcomes
across the store, and an asymmetric meta-governor defaults to the local advisor
unless pooling yields a strictly positive recent utility margin. We prove that
delayed outcomes escape the prompt-only regret floor, that neither fixed scope
suffices alone, and that switching completes within bounded time under bounded
feedback delay. Across LongMemEval, historical drift in TempLAMA and TAQA,
delays up to 96 episodes, and sparse or bandit feedback with Qwen3-30B-A3B and
GPT-4o, RegimeCouncil outperforms the stronger fixed scope in all 30 held-out
settings, gaining 5.2--7.3 points on disagreement episodes and +27.9 points
over standalone cold-start fallback under clustered shifts while reducing
harmful stale-memory exposures at microsecond-level overhead.
\end{abstract}

\section{Introduction}
\label{sec:intro}

Long-term memory enables language agents to personalize decisions across
extended interactions \citep{park2023generative,packer2023memgpt}, yet users'
circumstances, preferences, and external facts inevitably change over time.
Alongside deciding what to store and how to retrieve it, a fundamental question
in agent memory is how the agent should \emph{use} stored facts when they
conflict with an evolving reality
\citep{maharana2024locomo,wu2024longmemeval,hu2026memoryagentbench}.
Most existing frameworks treat conflict resolution as an explicit write-time
operation: once a timestamped correction or replacement value is provided, the
store overwrites or re-ranks the affected entry
\citep{chhikara2025mem0,rasmussen2025zep}.
In deployed assistants, however, facts routinely become stale silently without
manual deletion, and store updates may wait for asynchronous consolidation.
Conflicts therefore surface \emph{before} the store is refreshed: a delayed
complaint, thumbs-down, or failed task reveals that acting on a stored record
hurt, without yet providing the replacement value.
The key question is how to govern memory use in this \emph{pre-consolidation
window}.

\begin{figure}[t]
\centering
\includegraphics[width=\linewidth]{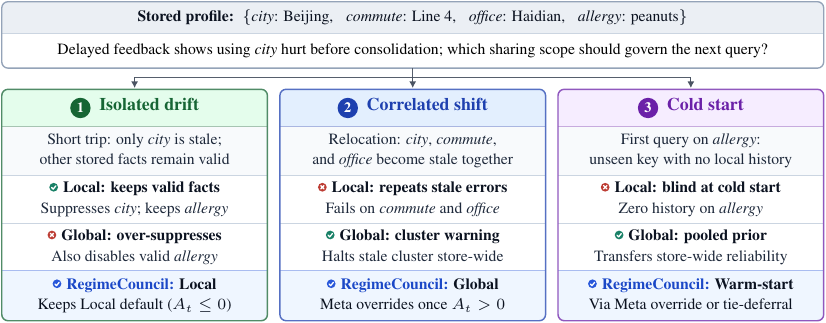}
\caption{Pre-consolidation memory conflict governance across three hidden
regimes: (1)~\emph{isolated drift}, where suppression should stay fact-local;
(2)~\emph{correlated shift}, where suppression should become store-wide; and
(3)~\emph{cold start}, where only pooled cross-key evidence is available.
RegimeCouncil defaults to Local, switches to Global when recent revealed
outcomes yield a positive counterfactual margin (\(A_t>0\);
Section~\ref{sec:meta}), and optionally warm-starts undecided keys via
tie-deferral.}
\label{fig:correction-scale}
\end{figure}

Consider an assistant storing user attributes such as \textit{city},
\textit{commute}, and \textit{allergy} (Figure~\ref{fig:correction-scale}).
A delayed complaint on \textit{city} reveals that using the stored record hurt,
but not which of three latent regimes holds:
under (1)~\emph{isolated drift}, only \textit{city} changed while other facts
remain valid, so store-wide suppression would discard useful memories;
under (2)~\emph{correlated shift} (e.g., relocation), related attributes shifted
together, so suppressing \textit{city} alone repeats stale-memory errors key by
key; and under (3)~\emph{cold start}, a newly queried attribute has no local
history, whereas pooled outcomes across other facts are already informative.
Feedback thus shows that a conflict occurred, but not whether suppression should
remain \emph{fact-local} or become \emph{store-wide}.

While recent memory architectures advance storage, compression, and retention
\citep{packer2023memgpt,chhikara2025mem0,rasmussen2025zep,zhou2026mem1,
zhang2026memrl,kang2026observability,kang2026retain}, they typically fix one
sharing scope at design time (Figure~\ref{fig:correction-scale}, Local and
Global rows).
A fact-local (\emph{local}) policy confines each outcome to the queried
attribute \citep{zhang2026memrl,chhikara2025mem0,rasmussen2025zep}, avoiding
cross-fact interference under isolated drift but failing under correlated shifts
and offering no guidance on unseen keys.
A store-wide (\emph{global}) policy pools recent outcomes via a shared gate or
recency window, halting correlated failures and warm-starting new keys at the
cost of letting a few volatile attributes disable stable memories.
Deciding online between these scopes---under hidden validity and delayed
feedback---is a sequential decision problem we term \emph{memory conflict
governance}.

To solve this problem without retraining the LLM or adding user-facing latency,
we propose \textbf{RegimeCouncil} (\textbf{RC}), a lightweight three-module
router.
First, an \emph{anchor-referenced scorer} isolates whether memory helped or hurt
on each revealed episode by comparing the memory-assisted answer against a
\emph{no-memory anchor}. This supports both off-path paired evaluation and
unbiased inverse-propensity estimation from executed-action-only bandit
feedback, so raw task difficulty does not masquerade as memory harm.
Second, two advisors operate at complementary scopes: a \emph{Local Advisor}
tracks only the queried fact and abstains conservatively on local ties to
minimize harmful exposure, while a \emph{Global Advisor} pools recent outcomes
across all stored facts. For long-delay cold-start coverage, a tie-deferral
variant, \textbf{RegimeCouncil-TD} (\textbf{RC-TD}), instead defers local ties
to global evidence.
Third, an asymmetric \emph{Meta Governor} follows the Local Advisor by default
and switches to the Global Advisor only when global pooling accumulates a
strictly positive counterfactual utility margin over recent revealed episodes:
Local is the safer default under isolated drift, whereas a Global override
requires strictly positive recent counterfactual gain.
Each module is backed by a formal guarantee: delayed outcome replay is necessary
to escape the prompt-only regret floor (Proposition~\ref{prop:A}); neither fixed
scope suffices alone, motivating dual advisors
(Proposition~\ref{prop:granularity}); and margin-gated arbitration switches
within bounded decisions under delay \(\tau\)
(Proposition~\ref{prop:meta-response}).

Our main contributions are threefold:
(1)~\textbf{Problem and architecture:} We formulate \emph{memory conflict
governance} and introduce \textbf{RegimeCouncil}, an \(O(1)\)-time online router
supporting both paired and bandit feedback at \(2.29\,\mu\mathrm{s}\) per
decision and zero added critical-path LLM calls.
(2)~\textbf{Theoretical guarantees:} We establish an observation-only regret
floor, linear-regret separations against both fixed scopes, and finite-time
switching bounds under delayed feedback
(Propositions~\ref{prop:A}--\ref{prop:meta-response}).
(3)~\textbf{Comprehensive evaluation:} Across LongMemEval update streams (16-
and 54-fact pools) and real historical chronologies on TempLAMA and TAQA with
Qwen3-30B-A3B and GPT-4o, RegimeCouncil outperforms the better fixed scope in
all \(30\) held-out settings (\(33{,}792\) episodes), maintains consistent gains
across multi-step delays (\(\tau\in[1,96]\)), sparse/noisy/bandit feedback, and
clustered shifts (\(+27.9\) points and \(59\%\) fewer harmful exposures than
standalone cold-start fallback); ablating any module reduces accuracy by
\(2\)--\(7\) points.

\section{Related Work}
\label{sec:related}

\paragraph{Long-term agent memory and knowledge updates.}
Agent memory systems extend context via tiered storage, retrieval, and temporal
knowledge graphs \citep{park2023generative,packer2023memgpt,chhikara2025mem0,rasmussen2025zep},
learned compression or item-level utility reweighting
\citep{zhou2026mem1,zhang2026memrl}, and budgeted retention
\citep{kang2026observability,kang2026retain}.
Benchmarks such as LoCoMo, LongMemEval, and MemoryAgentBench identify outdated
facts and knowledge updates as a primary failure mode over multi-session
dialogues \citep{maharana2024locomo,wu2024longmemeval,hu2026memoryagentbench}.
However, standard evaluations expose explicit timestamps in the prompt so a
recency-aware reader can identify the latest write directly, and existing memory
systems fix their feedback scope at design time---updating either the single
accessed record \citep{zhang2026memrl,chhikara2025mem0,rasmussen2025zep} or a
global retrieval gate.
We instead address silent staleness before store consolidation, dynamically
arbitrating between fact-local and store-wide feedback scopes online.

\paragraph{Non-stationarity and feedback sharing scope.}
Concept-drift detectors, non-stationary bandits, and strongly adaptive online
learners handle changing environments by discounting older observations along a
single stream \citep{gama2014survey,besbes2014nonstationary,cheung2019nonstationarity,
daniely2015strongly,jun2017coinbetting}.
Multi-fact agent memory induces structured non-stationarity across \(K\) keys:
facts may drift asynchronously (where cross-fact pooling causes interference) or
shift in correlated clusters (where pooling accelerates recovery).
Rather than tuning a single forgetting window, RegimeCouncil maintains both a
per-key window and a store-wide pooled window and selects the governing scope
online.

\paragraph{Expert advice and partial observability.}
Switching-expert algorithms track changing best predictors over time
\citep{herbster1998tracking,daniely2015strongly}, while POMDPs formalize
decision making when latent state is hidden from the current observation
\citep{krishnamurthy2016pac,jin2020sample}.
Generic expert-tracking rules treat experts symmetrically without exploiting
relational key structure or per-query baselines.
RegimeCouncil instead pairs memory against a no-memory anchor to cancel
query-difficulty variance (directly or via inverse-propensity bandit estimates)
and applies an asymmetric Meta Governor that defaults to the Local Advisor,
granting Global overrides only when cross-key pooling achieves strictly
positive recent counterfactual gain.

\section{Method}
\label{sec:method}

\begin{figure}[t]
\centering
\IfFileExists{figure2.png}{%
  \includegraphics[width=0.81\linewidth]{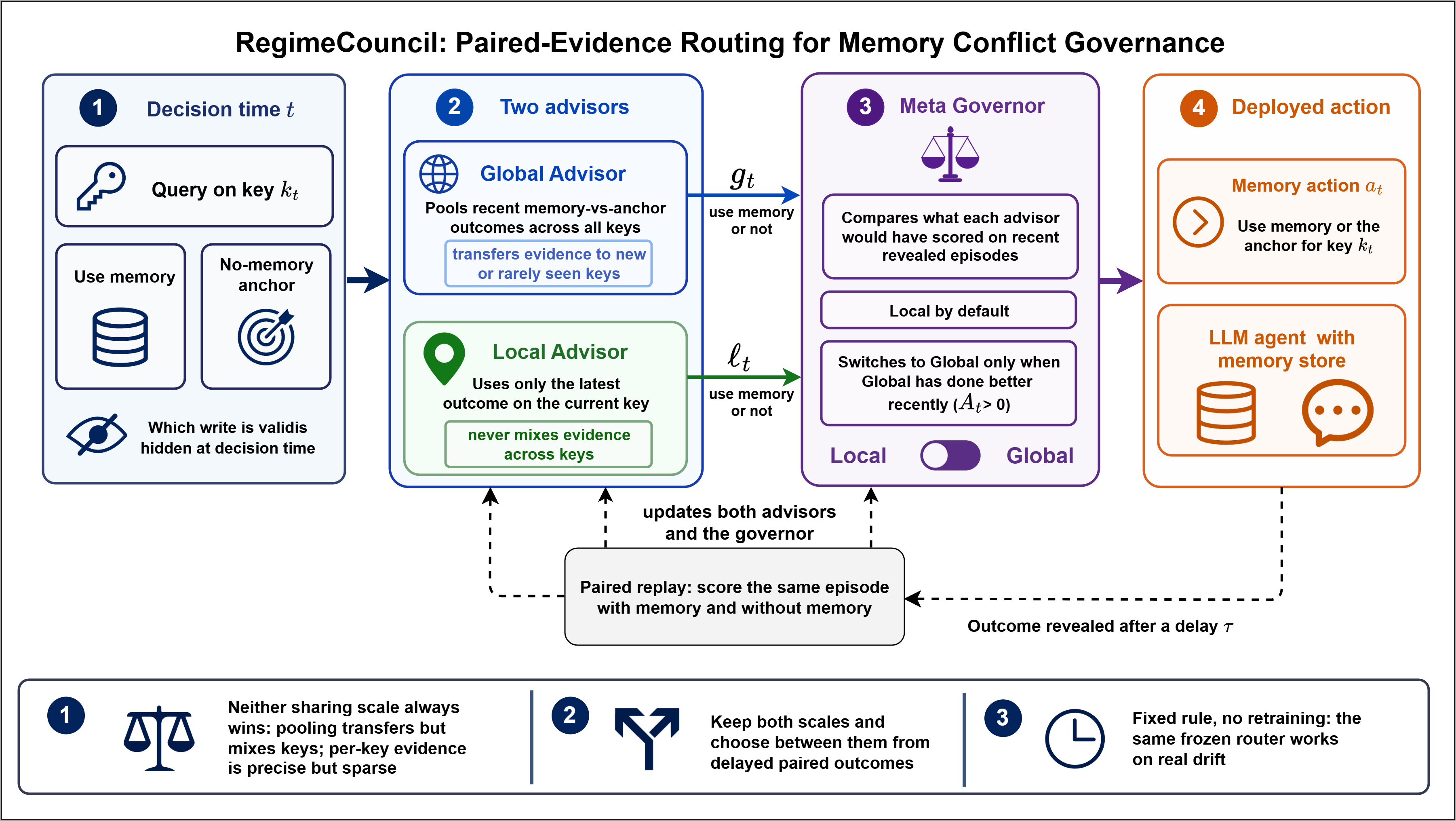}%
}{%
  \fbox{\parbox{0.9\linewidth}{\centering\vspace{4em}
  RegimeCouncil figure missing\vspace{4em}}}%
}
\caption{Overview of RegimeCouncil: (1)~query on key \(k_t\) with hidden
validity chooses memory (\(a_t{=}1\)) vs.\ no-memory anchor (\(a_t{=}0\));
(2)~Global pools recent outcomes across all keys while Local reads \(k_t\)'s
latest outcome; (3)~Meta Governor defaults to Local and switches to Global iff
\(A_t>0\); (4)~delayed feedback updates all three modules.}
\label{fig:regimecouncil}
\end{figure}

We first present the online decision setup (\S\ref{sec:setup}), our paired and
bandit evidence scorers (\S\ref{sec:paired}), and the three-module
RegimeCouncil router (\S\ref{sec:router}; Figure~\ref{fig:regimecouncil}),
followed by the formal guarantees that motivate each component
(\S\ref{sec:theory}).

\subsection{Problem setup}
\label{sec:setup}
An LLM agent maintains a long-term memory store of \(K\) tracked facts indexed
by keys \(k\in\{1,\dots,K\}\) (such as \textit{city}, \textit{commute}, or
\textit{allergy}).
At each sequential interaction \(t\in\{1,\dots,T\}\), a query arrives for key
\(k_t\in\{1,\dots,K\}\) with decision-time observation \(z_t\in\mathcal{Z}\)
(the user prompt together with the retrieved memory record), and the agent must
decide whether to inject that record under hidden validity.
Formally, whether the stored record for \(k_t\) is valid or stale is governed by
a latent state \(r_t\in\{0,1\}\) absent from \(z_t\) when facts change silently
(a \emph{collision}: the prompt matches the stored record under both states,
while the optimal memory action reverses).

At step \(t\), the router chooses a binary action \(a_t\in\mathcal{A}=\{0,1\}\),
where \(a_t{=}1\) injects the stored memory record for \(k_t\) and \(a_t{=}0\)
withholds it to answer with the base model alone (the \emph{no-memory anchor}),
receiving episode utility \(u_t\in[0,1]\).
Downstream feedback arrives with a delay of at most \(\tau\ge 1\) steps: prior
to choosing \(a_t\), the router observes history
\(h_t=(z_{1:t},a_{1:t-1},\mathcal{F}_t)\), where \(\mathcal{F}_t\) contains the
revealed outcomes of completed episodes \(s\le t-\tau\) (and any earlier
arrivals).
The objective is to maximize expected cumulative utility
\(J_T(P)=\mathbb{E}_P[\sum_{t=1}^T u_t]\) without retraining the LLM.

\subsection{Evidence scoring: paired anchor and bandit feedback}
\label{sec:paired}
Raw task rewards \(u_s\in[0,1]\) confound memory validity with intrinsic query
difficulty: \(u_s{=}0\) may simply mean the question was hard even without
memory, rather than that stored memory was harmful.
RegimeCouncil isolates the causal contribution of memory on each revealed
episode \(s\in\mathcal{F}_t\) via its \emph{paired utility difference}:
\[
d_s \;=\; u_{s,\mathrm{mem}}-u_{s,0}\;\in\;[-1,1],
\]
where \(u_{s,\mathrm{mem}}\in[0,1]\) is the utility with memory injected and
\(u_{s,0}\in[0,1]\) is the utility of the no-memory anchor on the same query.
Here \(d_s>0\) indicates memory helped, \(d_s<0\) indicates stale memory hurt,
and \(d_s=0\) is a tie.
Depending on the deployment environment, RegimeCouncil supports two feedback
modes with the exact same downstream routing architecture; both yield a scalar
paired difference \(d_s\) (or its unbiased estimate \(\hat d_s\)):
\begin{itemize}\setlength{\itemsep}{2pt}\setlength{\parskip}{0pt}\setlength{\topsep}{2pt}
\item \textbf{Asynchronous paired mode (zero critical-path latency):} The user
immediately receives the single response selected by \(a_s\). When delayed
feedback (e.g., a correction, complaint, or tool-execution check) arrives, an
off-the-critical-path background response without memory is evaluated against
that same feedback signal (or read from offline evaluation logs), yielding
\((u_{s,\mathrm{mem}},u_{s,0})\) directly with \(0\,\mathrm{ms}\) added
user-facing latency.
\item \textbf{Executed-action-only bandit mode (zero extra LLM calls):} When
downstream feedback reveals only the scalar outcome \(u_s\) of the deployed
action \(a_s\), mixing \(\varepsilon\)-exploration (\(\varepsilon{=}0.05\)) with
the router's proposal gives propensity \(p_s=\Pr(a_s{=}1\mid h_s)\in[\varepsilon,
1-\varepsilon]\) and the unbiased inverse-propensity weighting (IPW) estimator
\[
\hat d_s \;=\; \frac{\mathds{1}[a_s{=}1]\,u_s}{p_s}
             - \frac{\mathds{1}[a_s{=}0]\,u_s}{1-p_s},
\qquad \mathbb{E}[\hat d_s\mid h_s]=d_s,
\]
with single-arm utility estimates \(\hat u_{s,\mathrm{mem}}=\mathds{1}[a_s{=}1]u_s/p_s\)
and \(\hat u_{s,0}=\mathds{1}[a_s{=}0]u_s/(1-p_s)\) substituting directly into
the modules below.
\end{itemize}

\subsection{RegimeCouncil architecture}
\label{sec:router}
All three modules read from the same revealed stream \(\mathcal{F}_t\) using
event-indexed sliding windows, requiring only \(O(1)\) rolling sums per step.
The two advisors embody the store-wide and fact-local scopes, and the Meta
Governor arbitrates when they disagree.

\paragraph{Global Advisor.}
\phantomsection\label{sec:global}%
When a key is queried for the first time or a cluster of facts becomes stale
together, recent outcomes on other keys provide an immediate cross-key signal.
The \emph{Global Advisor} maintains a window \(\mathcal{W}_g(t)\) of the \(W_g\)
most recently revealed paired differences across \emph{all} keys in
\(\mathcal{F}_t\) and proposes memory iff their sum is strictly positive:
\[
g_t = \mathds{1}\left[\sum_{i\in\mathcal{W}_g(t)} d_i > 0\right].
\]
Global pooling warm-starts unseen keys (Figure~\ref{fig:correction-scale},
regime~3) and reacts rapidly to correlated shifts (regime~2), but mixes positive
and negative \(d_i\) when keys drift asynchronously.

\paragraph{Local Advisor and cold-start tie handling.}
\phantomsection\label{sec:local}%
To prevent cross-key interference, the \emph{Local Advisor} restricts attention
to the \(W_\ell\) most recently revealed outcomes \(\mathcal{W}_\ell(t;k_t)\) of
the currently queried key \(k_t\).
Because a key's validity is piecewise constant between updates, older outcomes
on \(k_t\) are more likely to predate a switch; we fix \(W_\ell{=}1\) by design
so \(\ell_t\) tracks the latest revealed outcome \(d_{\mathrm{last}}(k_t)\) on
key \(k_t\) (Figure~\ref{fig:correction-scale}, regime~1).
Whenever \(d_{\mathrm{last}}(k_t)>0\), \(\ell_t{=}1\); whenever
\(d_{\mathrm{last}}(k_t)<0\), \(\ell_t{=}0\).
When \(k_t\) has \emph{no decisive local signal}---at cold start
(\(\mathcal{W}_\ell(t;k_t){=}\emptyset\)) or when
\(d_{\mathrm{last}}(k_t){=}0\)---the conservative default abstains
(\(\ell_t{=}0\)), whereas a tie-deferral variant delegates undecided episodes to
the Global Advisor:
\[
\begin{aligned}
\ell_t &= \mathds{1}\!\left[\sum_{i\in\mathcal{W}_\ell(t;k_t)} d_i > 0\right]
&&(\textbf{RC},\text{ conservative tie-abstention}),\\
\ell_t^{\mathrm{TD}} &=
\begin{cases}
g_t & \text{if } \mathcal{W}_\ell(t;k_t){=}\emptyset \text{ or } d_{\mathrm{last}}(k_t){=}0,\\
\ell_t & \text{otherwise}
\end{cases}
&&(\textbf{RC-TD},\text{ cold-start tie-deferral}).
\end{aligned}
\]
Because the sum over an empty window is zero, the first rule automatically
yields \(\ell_t{=}0\) at cold start, never using memory on an undecided key
unless the Meta Governor overrides. Cold-start tie-deferral instead delegates
undecided local episodes to \(g_t\) while preserving per-key isolation whenever
\(d_{\mathrm{last}}(k_t)\neq 0\), trading a marginal increase in exposure for
higher cold-start coverage under long delays \(\tau\).

\paragraph{Meta Governor and online deployment rule.}
\phantomsection\label{sec:meta}%
Whenever \(g_t \neq \ell_t\), the \emph{Meta Governor} arbitrates by scoring
each advisor's counterfactual utility on recent revealed interactions
\(i\in\mathcal{F}_t\):
\[
v_{j,i}=
\begin{cases}
u_{i,\mathrm{mem}} & \text{if advisor } j\in\{g,\ell\} \text{ proposed } 1 \text{ on } i,\\[2pt]
u_{i,0} & \text{otherwise,}
\end{cases}
\qquad
A_t = \sum_{i\in\mathcal{W}_m(t)}\bigl(v_{g,i}-v_{\ell,i}\bigr),
\]
where \(\mathcal{W}_m(t)\) indexes the \(W_m\ge 1\) most recently revealed
episodes in \(\mathcal{F}_t\).
Because \(\mathcal{W}_m(t)\) is indexed by revealed events rather than
wall-clock steps, \(A_t\) handles variable delays and missing feedback
naturally.
Here \(A_t\) measures how much better Global has recently served revealed
episodes than Local; the governor applies an \emph{asymmetric} rule that keeps
the safer Local default unless Global demonstrates strictly positive recent net
gain (Figure~\ref{fig:correction-scale}, regime~2):
\[
a_t \;=\;
\begin{cases}
g_t & \text{if } A_t>0 \quad(\text{Global override}),\\[2pt]
\ell_t & \text{otherwise} \quad(\text{Local default}).
\end{cases}
\]
With \(W_\ell{=}1\) and override threshold \(0\) (\(A_t>0\)) fixed by design,
\((W_g,W_m)\) are the only hyperparameters: a wider pooled window \(W_g\)
averages cross-key evidence smoothly, whereas a shorter governor window \(W_m\)
tracks sustained shifts promptly.
The routing decision is deterministic apart from \(\varepsilon\)-exploration in
bandit mode, and uses only revealed outcomes, adding zero LLM calls or
critical-path latency.

\subsection{Theoretical guarantees}
\label{sec:theory}
We now state three formal guarantees---in the exact order of the modules
above---showing why each component of RegimeCouncil is necessary (complete
proofs in Appendix~\ref{app:proofs}).
We formalize silent staleness, bounded delay, and anchor evaluation as follows:

\begin{assumption}[Decision-time unobservability]
\label{ass:1}
On colliding contexts, the latent validity state \(r_t\) is not a function of
the decision-time observation \(z_t\).
\end{assumption}
\begin{assumption}[Delayed hindsight recoverability]
\label{ass:2}
An outcome generated at step \(s\) is revealed by step \(s+\tau\), and the
revealed comparison between using and withholding memory carries the sign of the
regime-optimal action on that episode.
\end{assumption}
\begin{assumption}[Anchor-referenced evaluation]
\label{ass:3}
For each revealed episode \(s\in\mathcal{F}_t\), the router observes
\((u_{s,\mathrm{mem}},u_{s,0})\) (or unbiased bandit estimates
\((\hat u_{s,\mathrm{mem}},\hat u_{s,0})\) under exploration propensity
\(p_s\in[\varepsilon,1-\varepsilon]\)).
\end{assumption}

Let \(u(a;z,r)=\mathbb{E}[u_t\mid a_t{=}a,z_t{=}z,r_t{=}r]\) denote mean utility
and \(\rho_r(a;z)=\max_{a'}u(a';z,r)-u(a;z,r)\ge 0\) its suboptimality gap.
A context \(z\) is \(\Delta\)-colliding if visited under both regimes with
\(\min_a[\rho_0(a;z)+\rho_1(a;z)]\ge\Delta\).
For a sequence \((z_t,r_t)\) with colliding-visit share \(p\) and minority-regime
share \(q\in(0,\tfrac12]\), we compare stationary observation-only policies
\(\Pi_{\mathrm{obs}}=\{\pi:a_t\sim\pi(\cdot\mid z_t)\}\) against history-aware
policies \(\Pi_{\mathrm{hist}}=\{\pi:a_t\sim\pi(\cdot\mid h_t)\}\).

\paragraph{1. Why track delayed outcome history? (Justifies the Evidence Scorer.)}
\begin{proposition}[Observation floor and history separation]
\label{prop:A}
\emph{(i) Floor.} Under Assumption~\ref{ass:1} and \(\Delta\)-collisions
\((p,q,\Delta)\), every observation-only policy \(\pi\in\Pi_{\mathrm{obs}}\)
incurs average regret \(\frac1T\mathrm{Regret}_T(\pi)\ge\delta:=pq\Delta\).
\emph{(ii) Separation.} If keys switch at most \(S\) times in total across \(K\)
keys under delay at most \(\tau\) (Assumption~\ref{ass:2}), replaying key
\(k_t\)'s latest revealed outcome (the Local Advisor, \S\ref{sec:local})
achieves \(\frac1T\mathrm{Regret}_T\le (S+K)m/T\), where \(m\) is the maximum
visits to a single key within any \(\tau\)-step window (e.g., \(m=\lceil\tau/K\rceil\)
under round-robin visits); hence \(\Pi_{\mathrm{hist}}\) strictly outperforms
\(\Pi_{\mathrm{obs}}\) once \(T>(S+K)m/\delta\).
\end{proposition}
\begin{corollary}[Model capacity cannot recover missing validity bits]
\label{cor:A1}
No stationary feature extractor \(f(z_t)\) followed by an arbitrary classifier
can reduce average regret below \(\delta\) when \(r_t\) is absent from \(z_t\).
\end{corollary}

\paragraph{2. Why retain both Local and Global Advisors? (Justifies Dual Advisors.)}
\begin{proposition}[Neither fixed scope is uniformly sufficient]
\label{prop:granularity}
There exist asynchronous-drift streams on which the Global Advisor incurs
\(\Omega(T)\) regret relative to the Local Advisor because majority-key outcomes
dominate \(\mathcal{W}_g\) and misclassify minority keys, and sparse-key streams
on which the Local Advisor incurs \(\Omega(T)\) regret relative to the Global
Advisor because unseen or rarely revisited keys lack local history while pooled
evidence is already informative.
\end{proposition}

\paragraph{3. Why does asymmetric gating adapt rapidly? (Justifies the Meta Governor.)}
\begin{proposition}[Finite-time switching under delayed feedback]
\label{prop:meta-response}
If \(v_{g,i}-v_{\ell,i}\ge\gamma>0\) holds for \(W_m\) consecutive revealed
episodes, the Meta Governor selects the Global Advisor; if
\(v_{g,i}-v_{\ell,i}\le -\gamma\) holds for \(W_m\) consecutive revealed
episodes, it selects the Local Advisor. Under feedback delay at most \(\tau\),
either switch completes within at most \(W_m+\tau\) decisions from the onset of
the streak.
\end{proposition}

\section{Experiments}
\label{sec:exp}
\setlength{\floatsep}{13pt plus 2pt minus 2pt}
\setlength{\textfloatsep}{14pt plus 2pt minus 2pt}

We evaluate RC and RC-TD (\S\ref{sec:local}) along four questions:
(1)~whether online scope arbitration outperforms the better fixed-scope advisor
on controlled update streams (\S\ref{sec:async-pareto});
(2)~whether the frozen router transfers to real-world historical drift
(\S\ref{sec:realdrift});
(3)~whether all three modules---and the Meta Governor beyond cold-start
fallback---are necessary across asynchronous and clustered regimes
(\S\ref{sec:module-ablation}); and
(4)~how the router behaves under multi-step delays, imperfect or bandit
feedback, and runtime constraints (\S\ref{sec:delay-bandit}).
Throughout all experiments, the Local Advisor uses a single-outcome window, the
switching threshold is fixed at zero, and the global and arbitration window
sizes (32 and 8) are selected once on disjoint development streams and held
frozen across every benchmark.
We test GPT-4o and Qwen3-30B-A3B (Qwen, about 3B active parameters), caching
completions per item once so all policies observe identical generations.

\subsection{Controlled knowledge-update streams}
\label{sec:async-pareto}

We construct multi-pass update streams from audited knowledge-update sessions in
LongMemEval \citep{wu2024longmemeval}.
Because raw benchmark dialogues contain explicit timestamps that reduce
staleness detection to surface recency reading rather than hidden validity
(Assumption~\ref{ass:1}), we remove explicit temporal cues with a deterministic
lexical filter and govern validity changes through a hidden schedule.
Our protocol separates two timescales: the store refresh interval (every \(6K\)
episodes, or 96 episodes on the 16-key pool, modeling batched consolidation) and
the outcome revelation delay \(\tau\) (fixed at \(\tau = 1\) in
Sections~\ref{sec:async-pareto}--\ref{sec:realdrift} to isolate consolidation
lag, then varied from 1 to 96 in \S\ref{sec:delay-bandit}).
In the primary Block-Switch family, two cohorts of eight keys flip in opposite
directions near mid-horizon across 22 passes (352 episodes), with Schedules~A
and~B using independent seeds for visit order and switch timing.
Because both advisors choose the same action whenever \(g_t = \ell_t\), routing
decisions differ only on the label-free disagreement slice (\(g_t \neq \ell_t\));
we therefore report disagreement-slice accuracy, its gain over the better
fixed-scope advisor, and full-stream accuracy.

\begin{table}[t]
\centering
\caption{Main results on Schedules~A and~B: disagreement-slice accuracy (\%) and
full-stream accuracy (\%) with key-cluster bootstrap 95\% CIs (5,000
replicates). Best G/L denotes the better fixed-scope advisor \(\max(G,L)\), and
Agree-only removes the Meta Governor (\S\ref{sec:module-ablation}). Jointly
resampling all four runs yields +5.1 [2.3, 8.1] on the disagreement slice and
+1.9 [0.9, 3.1] full-stream.}
\label{tab:rc-main-dis}
\par\kern4pt
\footnotesize
\renewcommand{\arraystretch}{1.10}
\setlength{\tabcolsep}{3.7pt}
\begin{tabular}{@{}llcccccccc@{}}
\toprule
 & & \multicolumn{5}{c}{Disagreement Slice (\(g_t \neq \ell_t\))} & \multicolumn{3}{c}{Full Stream (352 ep.)} \\
\cmidrule(lr){3-7}\cmidrule(lr){8-10}
Schedule & Model & $n_{\mathrm{dis}}$ & RC & Best G/L & Agree-only & Slice Gap & RC & Best G/L & Full Gap [95\% CI] \\
\midrule
Schedule~A & Qwen & 110 & \textbf{87.3} & 80.0 & 65.5 & \textbf{+7.3} & \textbf{70.2} & 67.9 & \textbf{+2.3} [0.0, 3.4] \\
 & GPT-4o & 134 & \textbf{84.3} & 79.1 & 67.2 & \textbf{+5.2} & \textbf{73.3} & 71.3 & \textbf{+2.0} [0.0, 3.4] \\
\addlinespace[2.5pt]
Schedule~B & Qwen & 151 & \textbf{83.4} & 78.1 & 76.2 & \textbf{+5.3} & \textbf{71.6} & 69.3 & \textbf{+2.3} [--0.3, 4.0] \\
 & GPT-4o & 146 & \textbf{82.9} & 77.4 & 76.0 & \textbf{+5.5} & \textbf{73.0} & 70.7 & \textbf{+2.3} [--0.3, 3.7] \\
\bottomrule
\end{tabular}
\end{table}

Table~\ref{tab:rc-main-dis} shows that RegimeCouncil outperforms \(\max(G,L)\)
by +5.2 to +7.3 disagreement-slice points (+2.0 to +2.3 full-stream), whereas
Agree-only---which drops the Meta Governor and uses memory only when both
advisors concur (\S\ref{sec:module-ablation})---trails at 65.5\%--76.2\%.
\textbf{Crucially, the better fixed scope itself reverses across streams}---Local
leads on Schedule~A and Global leads on Schedule~B for both LLMs---so any
design-time choice is suboptimal on one stream, exactly as
Proposition~\ref{prop:granularity} predicts, while RegimeCouncil adapts to both
without retuning (selecting an action at least as accurate as the alternative
advisor on 94\%--98\% of disagreements; worked examples in
Appendix~\ref{app:cases}).
Scaling up to the 54-key LongMemEval pool at matched refresh interval 324
(Table~\ref{tab:scale54}) yields +6.0 to +9.7 slice points, with all eight 95\%
bootstrap CIs above zero.
Across all 30 held-out settings (\(33{,}792\) episodes from 15 streams on both
LLMs: 16-key Block-Switch~A/B/C, Independent-Switch~A/B, and Poisson-Flip~A/B;
54-key Block-Switch~A/B and Independent-Switch~A/B; and TempLAMA~A/B and
TAQA~A/B; Appendix~\ref{app:protocol}), RegimeCouncil outperforms \(\max(G,L)\)
in every setting (+7.3 median slice points for RC and +9.8 for RC-TD).

\begin{table}[t]
\centering
\caption{Scale-up to the 54-key LongMemEval pool (refresh interval 324, frozen
windows): disagreement-slice gap over \(\max(G,L)\) (points) with key-bootstrap
95\% CIs (5,000 replicates). All eight intervals exclude zero.}
\label{tab:scale54}
\par\kern4pt
\footnotesize
\renewcommand{\arraystretch}{1.12}
\setlength{\tabcolsep}{7.8pt}
\begin{tabular}{@{}lcccc@{}}
\toprule
 & \multicolumn{2}{c}{Qwen3-30B-A3B} & \multicolumn{2}{c}{GPT-4o} \\
\cmidrule(lr){2-3}\cmidrule(lr){4-5}
Schedule Family & Schedule~A & Schedule~B & Schedule~A & Schedule~B \\
\midrule
Block-Switch & \textbf{+6.0} [0.6, 9.1] & \textbf{+8.4} [4.1, 10.0] & \textbf{+8.5} [3.0, 11.5] & \textbf{+9.0} [3.7, 10.6] \\
Independent-Switch & \textbf{+8.6} [5.1, 10.1] & \textbf{+9.2} [5.7, 10.7] & \textbf{+9.2} [4.6, 11.7] & \textbf{+9.7} [7.0, 12.4] \\
\bottomrule
\end{tabular}
\end{table}

\subsection{Transfer to real-world historical drift}
\label{sec:realdrift}
To evaluate naturally occurring drift beyond synthetic schedules, we test the
same frozen router on 120 evolving Wikidata facts from TempLAMA
\citep{dhingra2022time} (2010--2020) and 100 evolving Wikipedia-table questions
from TAQA \citep{zhao2024set} (2000--2023).
Each fact transitions at its true historical change year, with explicit dates
removed under Assumption~\ref{ass:1} and the store refresh interval matched at
\(6K\) episodes.
As shown in Table~\ref{tab:realmain}, RegimeCouncil attains 81.5\%--86.2\%
disagreement-slice accuracy, outperforming the better fixed-scope advisor by
+7.2 to +18.1 slice points (+3.0 to +5.4 full-stream points), with all eight
95\% confidence intervals strictly above zero.
On TempLAMA with GPT-4o, early concentrated switches turn Global's pooled
window negative and over-suppress keys that have not yet switched, while
one-step lag and cold-start ties leave fixed Local behind during the early
cluster (reducing both fixed scopes to \(\approx 66\%\)); the Meta Governor
overrides during the cluster and reverts to Local on still-valid keys,
maintaining 83.6\%--84.2\% (+17.5 to +18.1 points).

\begin{table}[t]
\centering
\caption{Real-world historical drift on TempLAMA (120 keys) and TAQA (100
keys): disagreement-slice accuracy (\%) under true historical switch years and
frozen windows. All eight 95\% bootstrap CIs exclude zero.}
\label{tab:realmain}
\par\kern4pt
\footnotesize
\renewcommand{\arraystretch}{1.10}
\setlength{\tabcolsep}{2.7pt}
\begin{tabular}{@{}llcccccc@{}}
\toprule
 & & & \multicolumn{5}{c}{Disagreement Slice (\(g_t \neq \ell_t\), Sched.\ A / B)} \\
\cmidrule(lr){4-8}
Benchmark & Model & Sched. & $n_{\mathrm{dis}}$ & RC & Best G/L & Slice Gap & 95\% CI \\
\midrule
TempLAMA & Qwen & A / B & 1105 / 1109 & \textbf{85.4 / 86.2} & 78.2 / 78.8 & \textbf{+7.2 / +7.4} & [5.8, 8.9] / [6.0, 9.1] \\
 & GPT-4o & A / B & 786 / 793 & \textbf{84.2 / 83.6} & 66.2 / 66.1 & \textbf{+18.1 / +17.5} & [15.2, 20.5] / [14.6, 19.8] \\
\addlinespace[2.5pt]
TAQA & Qwen & A / B & 856 / 848 & \textbf{83.4 / 84.0} & 74.6 / 75.0 & \textbf{+8.8 / +9.0} & [6.4, 10.8] / [7.0, 11.1] \\
 & GPT-4o & A / B & 702 / 682 & \textbf{81.5 / 82.4} & 69.8 / 70.1 & \textbf{+11.7 / +12.3} & [9.1, 14.7] / [9.5, 15.6] \\
\bottomrule
\end{tabular}
\end{table}

\subsection{Module ablations and necessity of the Meta Governor}
\label{sec:module-ablation}
\phantomsection\label{tab:secondary}%
Ablating any module of RegimeCouncil lowers accuracy across benchmarks and
models (Appendix~\ref{app:ablation}, Table~\ref{tab:module-ablation}).
On Schedule~A, removing the Global Advisor lowers full-stream accuracy by
2.0--2.3 points across Qwen and GPT-4o, while removing the Local Advisor
lowers accuracy by 3.7--6.0 points and more than doubles harmful
stale-memory exposures on GPT-4o (rising from 9 to 19 episodes).
Removing the Meta Governor reduces full-stream accuracy by 6.5--6.8
points (17.2--21.8 points on disagreements), with comparable drops of
3.0--7.0 points on TempLAMA and TAQA.

\begin{table}[t]
\centering
\caption{Arbitration comparison across Case~1 (30 held-out settings,
33,792 episodes: mean full-stream accuracy \%, median slice gap, and total
harmful episodes), Case~2 (54-key clustered shifts: slice accuracy \% and
harmful episodes in parentheses), and Case~1+2 (54-key two-stage stream:
asynchronous updates followed by store-wide flips).}
\label{tab:tie-meta-comp}
\par\kern4pt
\footnotesize
\setlength{\tabcolsep}{4.0pt}
\begin{tabular}{@{}lcccccc@{}}
\toprule
 & \multicolumn{3}{c}{Case~1: 30 Held-Out Streams} & \multicolumn{2}{c}{Case~2: Clustered Shifts} & Case~1+2 \\
\cmidrule(lr){2-4}\cmidrule(lr){5-6}\cmidrule(lr){7-7}
Routing Policy & Full & Gap & Total Harm & Sync-Multi & Sync-Switch & Two-Stage \\
\midrule
Follow-recent-winner & 71.7 & +8.0 & 746 (2.21\%) & 94.2 & 93.5 (37) & 69.5 (65) \\
Fixed Share (\(\eta = 2, \alpha = 0.3\)) & 71.7 & +8.0 & 746 (2.21\%) & 94.2 & 93.5 (37) & 66.8 (74) \\
CUSUM on surplus & 71.3 & +7.8 & 746 (2.21\%) & 93.1 & 92.8 (41) & 67.1 (70) \\
Local + tie-deferral (no Meta) & \textbf{72.4} & +9.7 & 752 (2.23\%) & 66.3 & 91.3 (91) & 69.7 (102) \\
\addlinespace[2pt]
\textbf{RC} (conservative) & 71.6 & +7.3 & \textbf{707 (2.09\%)} & \textbf{94.2} & \textbf{93.5 (37)} & 68.2 (75) \\
\textbf{RC-TD} (tie-deferral) & 72.3 & \textbf{+9.8} & 748 (2.21\%) & \textbf{94.2} & \textbf{93.8 (40)} & \textbf{71.2 (77)} \\
\bottomrule
\end{tabular}
\end{table}

We next ask whether the Meta Governor can be replaced by a static cold-start
fallback (``Local + tie-deferral, no Meta'') that consults Global only when Local
has no decisive signal and trusts Local otherwise.
Table~\ref{tab:tie-meta-comp} compares this heuristic against RegimeCouncil
across three evaluation settings: the 30 held-out streams (Case~1, where 81\% of
advisor disagreements stem from local ties; cf.\
Figure~\ref{fig:correction-scale}, regime~3), 54-key clustered shifts (Case~2;
cf.\ regime~2: Sync-Multi and Sync-Switch, where groups of facts become stale after
individual keys have already recorded positive local outcomes), and a Two-Stage
stream transitioning from asynchronous updates to store-wide drift.
Without the Meta Governor, cold-start fallback cannot override an outdated
positive local signal: accuracy on Sync-Multi collapses from 94.2\% to 66.3\%
(--27.9 points), and harmful exposures on Sync-Switch more than double (91 vs.\
37--40 episodes).
Across all three regimes (Table~\ref{tab:tie-meta-comp}), conservative RC
achieves the lowest total harmful exposure, whereas RC-TD pairs tie-deferral
with the Meta Governor to match 94.2\% on clustered shifts while leading on
Case~1 (+9.8 median slice gain) and Two-Stage streams (71.2\%).

\subsection{Multi-step delays, imperfect supervision, and runtime cost}
\label{sec:delay-bandit}

\begin{table}[t]
\centering
\caption{\textbf{Top (Panel~A):} Disagreement-slice gap vs.\ \(\max(G,L)\)
(points; positive values favor RC/RC-TD) across outcome delays \(\tau \in [1,
96]\) at fixed refresh interval \(6K\). \textbf{Bottom (Panel~B):} Robustness
across all 30 held-out settings under random delays, sparse/noisy/bandit
feedback, and runtime cost; paired entries \(a\,/\,b\) report the two listed
severity levels in order.}
\label{tab:delay-bandit-main}
\par\kern4pt
\footnotesize
\setlength{\tabcolsep}{4.6pt}
\begin{tabular}{@{}llcccccc@{}}
\toprule
\textbf{Panel A: Key Pool (Refresh \(= 6K\))} & Variant & \(\tau = 1\) & \(\tau = 4\) & \(\tau = 8\) & \(\tau = 16\) & \(\tau = 48\) & \(\tau = 96\) \\
\midrule
16-key pool (refresh 96) & RC & +5.8 & +3.6 & --1.2 & --4.9 & --11.1 & --24.4 \\
 & \textbf{RC-TD} & \textbf{+8.3} & \textbf{+7.7} & \textbf{+6.9} & \textbf{+5.9} & \textbf{+10.4} & \textbf{+6.2} \\
\addlinespace[2pt]
54-key pool (refresh 324) & RC & +8.6 & +7.7 & +6.8 & +4.8 & +0.6 & --7.3 \\
 & \textbf{RC-TD} & \textbf{+9.9} & \textbf{+9.6} & \textbf{+9.4} & \textbf{+9.4} & \textbf{+8.6} & \textbf{+7.5} \\
\addlinespace[2pt]
TempLAMA 120 / TAQA 100 & RC & +11.5 & +11.0 & +10.1 & +8.8 & +2.6 & 0.0 \\
 & \textbf{RC-TD} & \textbf{+12.3} & \textbf{+12.2} & \textbf{+11.8} & \textbf{+11.2} & \textbf{+9.4} & \textbf{+8.4} \\
\bottomrule
\end{tabular}
\par\vspace{7.5pt}
\setlength{\tabcolsep}{0pt}
\renewcommand{\arraystretch}{1.05}
\begin{tabular*}{\linewidth}{@{\extracolsep{\fill}}lll@{}}
\toprule
\textbf{Panel B: Supervision \& Cost} & \textbf{RegimeCouncil} & \textbf{Baseline Comparison} \\
\midrule
\rcTwo{Random outcome delay}{(up to 8 / 16 episodes)} &
  \rcTwo{\textbf{+6.7 / +5.9} median gap}{(RC)} &
  \rcTwo{Exceeds Fixed Share \& Follow}{(140/150 vs.\ 133/150 positive)} \\
\rcRowSep
\rcTwo{Sparse feedback}{(75\% / 50\% revealed)} &
  \rcTwo{\textbf{+5.7 / +3.2} (RC)}{\textbf{+9.8 / +8.7} (RC-TD)} &
  \rcTwo{Exceeds Follow \& Fixed Share}{(+5.5 / +2.9 median gap)} \\
\rcRowSep
\rcTwo{Noisy feedback}{(5\% / 10\% label flips)} &
  \rcTwo{\textbf{+3.5 / +1.0} median gap}{(RC)} &
  \rcTwo{Exceeds Follow \& Fixed Share}{(+3.1 / +0.7 median gap)} \\
\rcRowSep
\rcTwo{Executed-action bandit}{(IPW estimator \(\hat{d}_s\))} &
  \rcTwo{\textbf{68.9\%} full-stream accuracy (RC)}{\textbf{31.3} harmful episodes/stream} &
  \rcTwo{Exceeds Global \& Follow}{(63.2\%, 60.6 harm; 68.5\%)} \\
\rcRowSep
\rcTwo{Critical-path LLM calls}{\& local CPU routing time} &
  \rcTwo{\textbf{1 LLM call (+0\,ms latency)}}{\textbf{2.29\,\rcBoldMu s/decision} CPU time} &
  \rcTwo{440k decisions/s throughput}{(+1.41\,\textmu s over Global)} \\
\bottomrule
\end{tabular*}
\end{table}

Table~\ref{tab:delay-bandit-main} evaluates robustness under longer delays,
imperfect supervision, and runtime constraints with the same frozen windows
(details in Appendix~\ref{app:extended-analysis}).
In Panel~A, at refresh interval \(6K\), conservative RC follows the \(\tau/K\)
scaling of Proposition~\ref{prop:A}(ii): its gain stays positive up to
\(\tau = 48\) on the 54-key and 100--120-key pools, but turns negative on the
16-key pool once \(\tau \ge 8\) leaves most keys undecided.
Deferring undecided local keys to Global in RC-TD removes this bottleneck,
maintaining +5.9 to +12.3 slice points across all 126 pool--delay settings up
to \(\tau = 96\).
Varying the refresh interval independently (Table~\ref{tab:commit-sweep})
confirms that while always using memory suffices under single-pass write-back
(93.0\% accuracy), routing gains +0.5 to +5.6 full-stream points across lags of
2 to 12 passes.

In Panel~B, across all 30 held-out settings, RegimeCouncil outperforms Fixed
Share and Follow-recent-winner under random delays up to 16 episodes, 50\%
missing feedback, 5\%--10\% label flips, and mid-stream drift-rate shifts
(68.1\% accuracy, leading on all 16 streams).
Under executed-action bandit feedback (\S\ref{sec:paired}) with no extra LLM
calls, RegimeCouncil reaches 68.9\% accuracy, outperforming the Global policy
(63.2\%) in all 30 settings while halving harmful exposures (31.3 vs.\ 60.6 per
stream) and surpassing Follow-recent-winner in 23 of 30 settings, at zero
critical-path latency and 2.29\,\textmu s of CPU time per decision.

\section{Conclusion}
\label{sec:conclusion}
We studied \emph{memory conflict governance} in long-term LLM agents:
determining the scope over which delayed negative feedback should propagate when
stored facts become stale prior to consolidation. Within this pre-consolidation
window, no single fixed sharing scope is sufficient---fact-local isolation
misses cold-start transfer and correlated shifts, whereas store-wide pooling
allows volatile attributes to suppress valid stable ones. To resolve this
tension, we introduced RegimeCouncil, a lightweight router that evaluates each
memory-assisted response against a no-memory anchor, maintains complementary
Local and Global Advisors, and arbitrates between them via an asymmetric Meta
Governor that defaults to Local and switches to Global only when cross-key
pooling demonstrates a strictly positive recent counterfactual gain. The same
architecture supports both off-path paired evaluation and executed-action
bandit feedback, while its tie-deferral variant (RC-TD) sustains cold-start
coverage under long delays.

Theoretical analysis and experiments across LongMemEval, TempLAMA, and TAQA
demonstrate that RegimeCouncil outperforms the stronger fixed scope in all 30
held-out settings (+7.3 median disagreement-slice points for RC and +9.8 for
RC-TD) and recovers +27.9 points with 59\% fewer harmful stale-memory exposures
than standalone cold-start fallback under clustered shifts, requiring only
\(2.29\,\mu\mathrm{s}\) per decision and zero additional critical-path LLM
calls. More broadly, these findings indicate that as agents accumulate richer
memory stores over longer interaction horizons, the appropriate sharing scope
varies across regimes rather than remaining fixed, making online cross-scope
arbitration both practical and principled. Promising directions for future work
include multi-agent memory stores with overlapping, asynchronously updated
records and multi-step tool trajectories requiring credit assignment across
multiple retrieved memories from a terminal task outcome---settings in which
asymmetric default-to-Local governance provides a principled safeguard.

\bibliographystyle{iclr2027_conference}
\bibliography{refs}

\begin{thebibliography}{21}
\providecommand{\natexlab}[1]{#1}
\providecommand{\url}[1]{\texttt{#1}}
\expandafter\ifx\csname urlstyle\endcsname\relax
  \providecommand{\doi}[1]{doi: #1}\else
  \providecommand{\doi}{doi: \begingroup \urlstyle{rm}\Url}\fi

\bibitem[Besbes et~al.(2014)Besbes, Gur, and Zeevi]{besbes2014nonstationary}
Omar Besbes, Yonatan Gur, and Assaf Zeevi.
\newblock Stochastic multi-armed-bandit problem with non-stationary rewards.
\newblock In Z.~Ghahramani, M.~Welling, C.~Cortes, N.~Lawrence, and
  K.~Weinberger (eds.), \emph{Advances in Neural Information Processing
  Systems}, volume~27. Curran Associates, Inc., 2014.
\newblock URL
  \url{https://proceedings.neurips.cc/paper_files/paper/2014/file/91ba7292e5388b90b58d0b839a7f19ec-Paper.pdf}.

\bibitem[Cheung et~al.(2019)Cheung, Simchi-Levi, and
  Zhu]{cheung2019nonstationarity}
Wang~Chi Cheung, David Simchi-Levi, and Ruihao Zhu.
\newblock Learning to optimize under non-stationarity.
\newblock In \emph{Proceedings of the Twenty-Second International Conference on
  Artificial Intelligence and Statistics}, pp.\  1079--1087. PMLR, 2019.

\bibitem[Chhikara et~al.(2025)Chhikara, Khant, Aryan, Singh, and
  Yadav]{chhikara2025mem0}
Prateek Chhikara, Dev Khant, Saket Aryan, Taranjeet Singh, and Deshraj Yadav.
\newblock {Mem0}: Building production-ready {AI} agents with scalable long-term
  memory.
\newblock In \emph{{ECAI} 2025}, volume 413 of \emph{Frontiers in Artificial
  Intelligence and Applications}, pp.\  2993--3000. {IOS} Press, 2025.
\newblock \doi{10.3233/FAIA251160}.

\bibitem[Daniely et~al.(2015)Daniely, Gonen, and
  Shalev-Shwartz]{daniely2015strongly}
Amit Daniely, Alon Gonen, and Shai Shalev-Shwartz.
\newblock Strongly adaptive online learning.
\newblock In \emph{Proceedings of the 32nd International Conference on Machine
  Learning}, pp.\  1405--1411. PMLR, 2015.

\bibitem[Dhingra et~al.(2022)Dhingra, Cole, Eisenschlos, Gillick, Eisenstein,
  and Cohen]{dhingra2022time}
Bhuwan Dhingra, Jeremy~R. Cole, Julian~Martin Eisenschlos, Daniel Gillick,
  Jacob Eisenstein, and William~W. Cohen.
\newblock Time-aware language models as temporal knowledge bases.
\newblock \emph{Transactions of the Association for Computational Linguistics},
  10:\penalty0 257--273, 2022.

\bibitem[Gama et~al.(2014)Gama, {\v{Z}}liobait{\.e}, Bifet, Pechenizkiy, and
  Bouchachia]{gama2014survey}
Jo{\~a}o Gama, Indr{\.e} {\v{Z}}liobait{\.e}, Albert Bifet, Mykola Pechenizkiy,
  and Abdelhamid Bouchachia.
\newblock A survey on concept drift adaptation.
\newblock \emph{ACM Computing Surveys}, 46\penalty0 (4), 2014.

\bibitem[Herbster \& Warmuth(1998)Herbster and Warmuth]{herbster1998tracking}
Mark Herbster and Manfred~K. Warmuth.
\newblock Tracking the best expert.
\newblock \emph{Machine Learning}, 32\penalty0 (2):\penalty0 151--178, 1998.
\newblock \doi{10.1023/A:1007424614876}.

\bibitem[Hu et~al.(2026)Hu, Wang, and McAuley]{hu2026memoryagentbench}
Yuanzhe Hu, Yu~Wang, and Julian McAuley.
\newblock Evaluating memory in {LLM} agents via incremental multi-turn
  interactions.
\newblock In \emph{International Conference on Learning Representations}, 2026.

\bibitem[Jin et~al.(2020)Jin, Kakade, Krishnamurthy, and Liu]{jin2020sample}
Chi Jin, Sham Kakade, Akshay Krishnamurthy, and Qinghua Liu.
\newblock Sample-efficient reinforcement learning of undercomplete {POMDP}s.
\newblock In \emph{NeurIPS}, 2020.

\bibitem[Jun et~al.(2017)Jun, Orabona, Wright, and Willett]{jun2017coinbetting}
Kwang-Sung Jun, Francesco Orabona, Stephen Wright, and Rebecca Willett.
\newblock Improved strongly adaptive online learning using coin betting.
\newblock In \emph{Proceedings of the 20th International Conference on
  Artificial Intelligence and Statistics}, pp.\  943--951. PMLR, 2017.

\bibitem[Kang et~al.(2026{\natexlab{a}})Kang, Liu, Kai, Liang, Tang, Cui,
  Zhong, and Yuan]{kang2026retain}
Qingcan Kang, Mingyang Liu, Shixiong Kai, Kaichao Liang, Zhentao Tang, Yuqi
  Cui, Tao Zhong, and Mingxuan Yuan.
\newblock Retain or consolidate? budget-dependent operator selection for
  language agent memory.
\newblock \emph{arXiv preprint arXiv:2607.17545}, 2026{\natexlab{a}}.

\bibitem[Kang et~al.(2026{\natexlab{b}})Kang, Liu, Kai, Liang, Zhong, and
  Yuan]{kang2026observability}
Qingcan Kang, Mingyang Liu, Shixiong Kai, Kaichao Liang, Tao Zhong, and
  Mingxuan Yuan.
\newblock Learning what to remember: Observability-safe memory retention via
  constrained optimization for long-horizon language agents.
\newblock \emph{arXiv preprint arXiv:2606.10616}, 2026{\natexlab{b}}.

\bibitem[Krishnamurthy et~al.(2016)Krishnamurthy, Agarwal, and
  Langford]{krishnamurthy2016pac}
Akshay Krishnamurthy, Alekh Agarwal, and John Langford.
\newblock Pac reinforcement learning with rich observations.
\newblock In \emph{NeurIPS}, 2016.

\bibitem[Maharana et~al.(2024)Maharana, Lee, Tulyakov, Bansal, Barbieri, and
  Fang]{maharana2024locomo}
Adyasha Maharana, Dong-Ho Lee, Sergey Tulyakov, Mohit Bansal, Francesco
  Barbieri, and Yuwei Fang.
\newblock Evaluating very long-term conversational memory of {LLM} agents.
\newblock In \emph{Proceedings of the 62nd Annual Meeting of the Association
  for Computational Linguistics (Volume 1: Long Papers)}, pp.\  13851--13870,
  2024.
\newblock \doi{10.18653/v1/2024.acl-long.747}.

\bibitem[Packer et~al.(2023)Packer, Wooders, Lin, Fang, Patil, Stoica, and
  Gonzalez]{packer2023memgpt}
Charles Packer, Sarah Wooders, Kevin Lin, Vivian Fang, Shishir~G. Patil, Ion
  Stoica, and Joseph~E. Gonzalez.
\newblock {MemGPT}: Towards {LLM}s as operating systems.
\newblock \emph{arXiv preprint arXiv:2310.08560}, 2023.
\newblock \doi{10.48550/arXiv.2310.08560}.

\bibitem[Park et~al.(2023)Park, O'Brien, Cai, Morris, Liang, and
  Bernstein]{park2023generative}
Joon~Sung Park, Joseph~C. O'Brien, Carrie~J. Cai, Meredith~Ringel Morris, Percy
  Liang, and Michael~S. Bernstein.
\newblock Generative agents: Interactive simulacra of human behavior.
\newblock In \emph{Proceedings of the 36th Annual ACM Symposium on User
  Interface Software and Technology}, 2023.
\newblock \doi{10.1145/3586183.3606763}.

\bibitem[Rasmussen et~al.(2025)Rasmussen, Paliychuk, Beauvais, Ryan, and
  Chalef]{rasmussen2025zep}
Preston Rasmussen, Pavlo Paliychuk, Travis Beauvais, Jack Ryan, and Daniel
  Chalef.
\newblock {Zep}: A temporal knowledge graph architecture for agent memory.
\newblock \emph{arXiv preprint arXiv:2501.13956}, 2025.
\newblock \doi{10.48550/arXiv.2501.13956}.

\bibitem[Wu et~al.(2025)Wu, Wang, Yu, Zhang, Chang, and Yu]{wu2024longmemeval}
Di~Wu, Hongwei Wang, Wenhao Yu, Yuwei Zhang, Kai-Wei Chang, and Dong Yu.
\newblock {LongMemEval}: Benchmarking chat assistants on long-term interactive
  memory.
\newblock In \emph{International Conference on Learning Representations}, 2025.
\newblock URL \url{https://openreview.net/forum?id=pZiyCaVuti}.

\bibitem[Zhang et~al.(2026)Zhang, Wang, Zhou, Liao, Feng, Li, Zheng, Zhang,
  Wen, Li, Xiong, Qi, Tang, and Wen]{zhang2026memrl}
Shengtao Zhang, Jiaqian Wang, Ruiwen Zhou, Junwei Liao, Yuchen Feng, Zhuo Li,
  Yujie Zheng, Weinan Zhang, Ying Wen, Zhiyu Li, Feiyu Xiong, Yutao Qi,
  Bo~Tang, and Muning Wen.
\newblock {MemRL}: Self-evolving agents via runtime reinforcement learning on
  episodic memory.
\newblock \emph{arXiv preprint arXiv:2601.03192}, 2026.

\bibitem[Zhao et~al.(2024)Zhao, Brumbaugh, Wang, Hajishirzi, and
  Smith]{zhao2024set}
Bowen Zhao, Zander Brumbaugh, Yizhong Wang, Hannaneh Hajishirzi, and Noah~A.
  Smith.
\newblock Set the clock: Temporal alignment of pretrained language models.
\newblock In \emph{Findings of the Association for Computational Linguistics:
  ACL 2024}, pp.\  15015--15040, 2024.
\newblock \doi{10.18653/v1/2024.findings-acl.892}.

\bibitem[Zhou et~al.(2026)Zhou, Qu, Wu, Kim, Prakash, Rus, Low, and
  Liang]{zhou2026mem1}
Zijian Zhou, Ao~Qu, Zhaoxuan Wu, Sunghwan Kim, Alok Prakash, Daniela Rus, Bryan
  Kian~Hsiang Low, and Paul Liang.
\newblock Mem1: Learning to synergize memory and reasoning for efficient
  long-horizon agents.
\newblock In \emph{International Conference on Learning Representations},
  volume 2026, pp.\  58413--58438, 2026.

\end{thebibliography}

\appendix
\setlength{\floatsep}{12pt plus 2pt minus 2pt}
\setlength{\textfloatsep}{13pt plus 2pt minus 2pt}

\phantomsection
\section*{Appendix overview}
\label{app:overview}
The appendices support the main text as follows.
Appendix~\ref{app:proofs} gives complete proofs for the three propositions in
\S\ref{sec:theory}, in the same order as the modules they justify.
Appendix~\ref{app:protocol} documents the evaluation protocol---item
selection, hidden schedules, cached completions, the revelation slice, module
ablations, sensitivity grids, scale-up, real-drift construction, the
accuracy--harm frontier, and the supplementary observable-conflict check---and
collects the per-setting tables summarized in \S\ref{sec:exp}.
Appendix~\ref{app:prompts} lists the fixed prompt templates; prompt wording is
shared across all routing policies, so it cannot favor one policy over another.
Appendix~\ref{app:cases} traces two worked advisor-disagreement episodes.
Appendix~\ref{app:extended-analysis} reports the full store-refresh sweep,
mid-conversation drift-rate shifts, and executed-action bandit and runtime
profiling that support \S\ref{sec:module-ablation}--\S\ref{sec:delay-bandit}.

\section{Proofs}
\label{app:proofs}

Section~\ref{sec:theory} states three theoretical results supporting each module
of RegimeCouncil: Proposition~\ref{prop:A} (an observation-only regret floor and
per-key history separation), Proposition~\ref{prop:granularity} (why neither a
purely global nor a purely local scope suffices alone), and
Proposition~\ref{prop:meta-response} (finite-time margin-gated switching under
delayed feedback). Each proof below is preceded by a one-sentence intuition
stating what it establishes.

\subsection{Observation floor and history separation (Proposition~\ref{prop:A})}
\label{app:prop-a}

\emph{Intuition.} Any policy that sees only the current prompt must play the
same action distribution on every visit to a colliding context, regardless of
the hidden regime; a history-aware policy can instead replay the latest
revealed outcome and pay regret only on visits that precede the first
post-switch outcome.

\begin{proof}[Proof of Proposition~\ref{prop:A}]
Fix a colliding context \(z\) visited \(n(z)\) times, of which a
visit-weighted minority-regime share \(q(z)=\min\{n_0(z),n_1(z)\}/n(z)\)
occurs under the minority regime. For every action \(a\),
\[
\rho_0(a;z)+\rho_1(a;z)\ge\Delta
\]
by \(\Delta\)-collision, and both \(\rho_r\) are nonnegative, so
\[
n_0(z)\,\rho_0(a;z)+n_1(z)\,\rho_1(a;z)
\;\ge\;
\min\{n_0(z),n_1(z)\}\bigl[\rho_0(a;z)+\rho_1(a;z)\bigr]
\;\ge\;
n(z)\,q(z)\,\Delta.
\]
Because \(\pi\in\Pi_{\mathrm{obs}}\) conditions only on \(z\), the same action
law \(\pi(\cdot\mid z)\) governs every visit to \(z\) irrespective of the
regime (Assumption~\ref{ass:1}), so the expected total regret accumulated at
\(z\) equals
\(\mathbb{E}_{a\sim\pi(\cdot\mid z)}\!\bigl[n_0(z)\rho_0(a;z)+n_1(z)\rho_1(a;z)\bigr]\).
The displayed bound holds for every \(a\), so taking the expectation preserves
it: the expected regret at \(z\) is at least \(n(z)q(z)\Delta\), and
randomization cannot undercut it. Summing over colliding contexts
(non-colliding visits contribute nonnegative regret) and dividing by \(T\)
yields average regret at least
\[
\frac{\Delta}{T}\sum_{z} n(z)\,q(z)
\;=\; p\,q\,\Delta
\;=\;\delta,
\]
where \(p\) is the colliding-visit frequency and \(q\) the visit-weighted
minority share. No independence across time is used. This proves part~(i).

\emph{Part (ii): separation.}
Consider the history policy that, on each key \(k\), replays the sign of the
revealed paired outcome for \(k\) with the latest generation time and defaults to the anchor
before any outcome for \(k\) is revealed---the conservative Local Advisor of
Section~\ref{sec:local} with \(W_\ell=1\), which conditions on \(h_t\) and so
lies in \(\Pi_{\mathrm{hist}}\). Fix a key \(k\) whose regime is constant
between switches; the regime-optimal memory action on \(k\) is determined by
that regime, and by Assumption~\ref{ass:2} each revealed comparison carries its
sign. Once a single post-switch outcome for \(k\) has been
revealed, every later-generated outcome carries the same sign, so the policy
plays the regime-optimal memory action on every subsequent visit until the
next switch and incurs zero regret there. Regret can therefore accrue only on
visits that precede the first revealed outcome after a switch, or before the
first outcome ever at cold start. An outcome generated at step \(s\) is
revealed by \(s+\tau\), so the stale visits following one switch number at most
\(m\), the largest count of visits to a single key within any window of
\(\tau\) steps. With at most \(S\) switches and \(K\) cold starts, and per-step
regret at most \(1\) (utilities lie in \([0,1]\)), the total regret is at most
\((S+K)m\); dividing by \(T\) gives the stated bound, which falls below the
floor \(\delta\) of part~(i) once \(T>(S+K)m/\delta\).

\emph{Complementary sparse-key regime.}
When keys are too sparse for per-key replay (each seen \(O(1)\) times), the
complementary history policy---the Global Advisor pooling across keys---attains
the same separation whenever the keys share a regime: after each of the \(S\)
switches its pooled window is corrected within \(O(W_g+\tau)\) revealed
outcomes, giving total regret \(O(S(W_g+\tau))\) and average regret below
\(\delta\) for large \(T\). The two cases mirror the constructions of
Proposition~\ref{prop:granularity}.
\end{proof}

\begin{proof}[Proof of Corollary~\ref{cor:A1}]
Any stationary feature map \(f(z)\) composed with a classifier induces a
stationary policy in \(\Pi_{\mathrm{obs}}\). Apply Proposition~\ref{prop:A}.
The floor is therefore invariant in the capacity of \(f\): enlarging a class
of maps \(\mathcal{Z}\to\Delta(\mathcal{A})\) cannot recover a regime bit that
is absent from \(z\).
\end{proof}

\subsection{Granularity constructions (Proposition~\ref{prop:granularity})}
\label{app:prop-granularity}

\emph{Intuition.} The two constructions are disjoint failure modes: the first
makes pooling biased by a majority key, the second makes isolation
uninformative by sparse visits. No single fixed scope can avoid both.

\begin{proof}[Proof of Proposition~\ref{prop:granularity}]
\emph{Global mixing.}
Use two keys with \emph{opposite} per-key advantages---\(d_t=+1\) whenever the
majority key \(k_1\) is queried and \(d_t=-1\) whenever the minority key
\(k_2\) is queried---and interleave them on a fixed periodic schedule in which
\(k_2\) appears a fraction \(\beta\in(0,\tfrac12)\) of the steps. Once the
window \(\mathcal{W}_g\) is full, it contains at most \(\lceil\beta W_g\rceil\)
entries equal to \(-1\), so its pooled sum is at least
\(W_g(1-2\beta)-O(1)>0\) for \(W_g\) large enough; the Global Advisor therefore
proposes memory on every step and acts wrongly on every \(k_2\) visit, a
fraction \(\beta\) of the \(T\) steps. The Local Advisor reads only the current key and, after a
single revealed visit, tracks the correct per-key sign on both keys. The
per-step utility gap on \(k_2\) visits is bounded below by a positive constant
on \(\Theta(T)\) steps, so Global suffers linear loss relative to Local.

\emph{Local sparsity.}
Present a sequence of distinct keys, each visited once, all sharing a common
positive memory advantage \(d_t=+1\). On a fresh key the conservative Local
Advisor has an empty per-key window and defaults to the anchor, incurring a
constant per-step gap. Once the first outcome is revealed, within \(\tau\)
steps, the Global Advisor's pooled sum is strictly positive and transfers the
shared advantage to every subsequent new key. Repeating over \(T\) distinct keys
yields linear loss for Local relative to Global.

Combining both streams establishes Proposition~\ref{prop:granularity}.
\end{proof}

\subsection{Finite meta response (Proposition~\ref{prop:meta-response})}
\label{app:prop-meta}

\emph{Intuition.} A persistent margin of at least \(\gamma\) fills the
\(W_m\)-window within \(W_m+\tau\) decisions, so the sign of \(A_t\) and the
selected advisor are determined in bounded time.

\begin{proof}[Proof of Proposition~\ref{prop:meta-response}]
Recall \(A_t=\sum_{i\in\mathcal{W}_m(t)}(v_{g,i}-v_{\ell,i})\) sums the surplus
of the \(W_m\) most recently \emph{revealed} episodes.
Suppose first that \(v_{g,i}-v_{\ell,i}\ge\gamma>0\) for the \(W_m\) consecutive
revealed indices that occupy the window. Then
\(A_t\ge W_m\gamma>0\), so the governor selects Global under the rule
``override iff \(A_t>0\).''
If instead \(v_{g,i}-v_{\ell,i}\le-\gamma\) throughout the window, then
\(A_t\le -W_m\gamma<0\) and Local remains (or returns as) the selection.

It remains to account for delay. An outcome generated at step \(s\) is
revealed by step \(s+\tau\). Filling \(\mathcal{W}_m(t)\) with \(W_m\) revealed
outcomes therefore requires at most \(W_m\) generations plus \(\tau\)
waiting steps, i.e., at most \(W_m+\tau\) decisions. The bound is
deterministic under the persistent-margin hypothesis; it does not cover
streams where the sign of \(v_{g,i}-v_{\ell,i}\) alternates inside every window
of length \(W_m\), nor single-trajectory variance of inverse-propensity bandit
estimates (which we evaluate empirically across all $30$ held-out settings in
\S\ref{sec:delay-bandit} and Appendix~\ref{app:feedback-cost}).
Because \(\mathcal{W}_m(t)\) is ordered by generation step, a late-arriving
outcome never displaces newer ones. On episodes where both advisors agree, the
margin is \(0\); the same argument shows that Global is selected whenever the
margins in the window are nonnegative and at least one is strictly positive,
and Local whenever they are all nonpositive.
\end{proof}

\section{Evaluation protocol and supplementary tables}
\label{app:protocol}

This appendix documents how the delayed-feedback streams are constructed, how
completions are scored, and how performance breaks down across individual
settings referenced in Section~\ref{sec:exp}. Each subsection begins with the
question it answers; tables marked \emph{report-only} were never used to tune
the frozen \((W_g, W_\ell, W_m, \text{gate}) = (32, 1, 8, 0)\) configuration.

\subsection{Items and cached answers}
\label{app:items}
\emph{Which items are used, and how are completions scored?}
The primary confirmatory set consists of 16 LongMemEval knowledge-update
questions \citep{wu2024longmemeval} whose initial and updated answers were
manually audited. Items were selected deterministically via SHA-256 hash from
audited items containing temporal cues \emph{before} any router evaluation, so
the confirmatory set could not be influenced by routing outcomes; the full
54-key audited pool is evaluated in Table~\ref{tab:scale54} and
Appendix~\ref{app:scale}.
To enforce decision-time unobservability (Assumption~\ref{ass:1}), we apply a
deterministic lexical filter that strips explicit temporal pointers (such as
currently, now, previously, and no longer) identically across all items without
altering facts or adding distractors.
Each key thus presents a colliding observation consisting of two unordered
memory records and a question with no recency cue.

For each key and model, we cache three completions once: answering without
memory (the no-memory anchor), answering with the old snapshot, and answering
with the updated snapshot.
At episode \(t\), \(u_{t,0}\) scores the no-memory answer against the currently
valid ground-truth answer, and \(u_{t,\mathrm{mem}}\) scores the answer
produced from the currently committed snapshot (falling back to \(u_{t,0}\)
before the first snapshot commit).
A completion scores 1 if normalized strings match (exact or substring match)
and 0 otherwise.
Caching ensures that every routing baseline observes identical LLM outputs on
every episode, isolating routing decisions from generation variance.

\subsection{Hidden schedule and delay}
\label{app:schedule}
\emph{How are validity changes scheduled, and where does delay enter?}
In the primary protocol, each of the 16 keys is visited once per pass over 22
passes (\(T = 352\) episodes).
Under the Block-Switch family, two cohorts of eight keys switch in opposite
directions near mid-horizon (staggered by 1--2 visits per key), while visit
order within each pass is deterministically hashed from the schedule seed.
Memory snapshots are committed every 96 episodes (6 passes), so the stored
record can lag behind a key's switch by up to 96 episodes.
In Sections~\ref{sec:async-pareto}--\ref{sec:realdrift}, completed outcomes are
revealed before the next step (\(\tau = 1\) in Assumption~\ref{ass:2}) to
isolate snapshot consolidation lag cleanly; Section~\ref{sec:delay-bandit} and
Appendix~\ref{app:delay-commit} vary the commit interval and outcome delay
\(\tau \in [1, 96]\) independently.
Window sizes \((W_g, W_m) = (32, 8)\) are selected once on disjoint development
schedules and held frozen across all held-out evaluations (Schedules~A, B,
and~C, plus a dedicated harm-exposure schedule in Appendix~\ref{app:frontier}).
Table~\ref{tab:commit-sweep} in Appendix~\ref{app:delay-commit} varies the
refresh interval at fixed \(\tau=1\); Panel~A of
Table~\ref{tab:delay-bandit-main} varies \(\tau\) at fixed refresh \(6K\).
The two scans are complementary and should not be conflated.

\subsection{Revelation slice}
\label{app:reveal}
\emph{Where does the routing gain actually come from?}
Table~\ref{tab:reveal} splits Schedules~A and~B at the first snapshot commit.
Until episode 96, any key without a committed snapshot is scored with the
no-memory utility (Appendix~\ref{app:items}), so the paired difference is zero,
Global and Local propose the same action, and RegimeCouncil matches both.
All 110, 134, 151, and 146 disagreement episodes (Schedule~A with Qwen and
GPT-4o, followed by Schedule~B with Qwen and GPT-4o)---and therefore the entire
routing gain in Table~\ref{tab:rc-main-dis}---occur after the first commit.
In most of these episodes (84, 99, 134, and 132), the key's latest paired
outcome is a tie, so conservative Local abstains while Global transfers
evidence from other keys, illustrating the local-sparsity regime of
Proposition~\ref{prop:granularity} (cf.\ Figure~\ref{fig:correction-scale},
regime~3); in the remaining episodes, the key's latest outcome disagrees with
the pooled window, illustrating the global-mixing regime (cf.\ regime~1).

\begin{table}[h]
\centering
\caption{Pre-commit vs.\ post-commit breakdown on the cached main streams.
Before episode 96 (Pre-Commit), Global, Local, and RegimeCouncil coincide
(\(n_{\mathrm{dis}} = 0\)). Post-Commit Slice Gap reports
\(\mathrm{RC} - \max(G,L)\) (points) on advisor disagreements after the first
store refresh, matching Table~\ref{tab:rc-main-dis}. Positive values indicate
RegimeCouncil outperforms \(\max(G,L)\).}
\label{tab:reveal}
\par\kern4pt
\footnotesize
\renewcommand{\arraystretch}{1.10}
\setlength{\tabcolsep}{6.5pt}
\begin{tabular}{@{}llccccc@{}}
\toprule
 & & \multicolumn{2}{c}{Pre-Commit (ep.\ 1--96)} & \multicolumn{3}{c}{Post-Commit (ep.\ 97--352)} \\
\cmidrule(lr){3-4}\cmidrule(lr){5-7}
Schedule & Model & Episodes & Accuracy (\%) & Episodes & $n_{\mathrm{dis}}$ & Slice Gap \\
\midrule
Schedule~A & Qwen & 96 & 37.5 & 256 & 110 & \textbf{+7.3} \\
 & GPT-4o & 96 & 43.8 & 256 & 134 & \textbf{+5.2} \\
\addlinespace[2.5pt]
Schedule~B & Qwen & 96 & 37.5 & 256 & 151 & \textbf{+5.3} \\
 & GPT-4o & 96 & 43.8 & 256 & 146 & \textbf{+5.5} \\
\bottomrule
\end{tabular}
\end{table}

\subsection{Module ablation}
\label{app:ablation}
\emph{Which component contributes what?}
Table~\ref{tab:module-ablation} reports the complete per-schedule module
ablation summarized in Section~\ref{sec:module-ablation} across Schedules~A
and~B.
On Schedule~B, removing any component lowers full-stream accuracy by 2.3 to 3.1
points.
Removing Global or Local leaves the remaining advisor to act alone, whereas
removing the Meta Governor (\S\ref{sec:meta}) yields the Agree-only variant,
which uses memory only when both advisors concur (\(g_t = \ell_t = 1\)).
The table reports both \emph{accuracy loss} relative to Full RC and
\emph{harmful exposure} (stale-memory episodes), showing that dropping Global
or Meta primarily hurts accuracy whereas dropping Local also inflates harmful
exposure.

\begin{table}[h]
\centering
\caption{Full module ablation on Schedules~A and~B (full-stream accuracy, \%).
Without the Meta Governor (\S\ref{sec:meta}), accuracy falls below either
fixed-scope advisor alone in all four runs (on Schedule~A, --6.8 points on Qwen
and --6.5 on GPT-4o; --21.8 and --17.2 points on disagreements), while removing
the Local Advisor roughly doubles harmful stale-memory exposures. Accuracy
changes are computed from unrounded values; negative $\Delta$ means lower
accuracy than Full RC.}
\label{tab:module-ablation}
\par\kern4pt
\footnotesize
\renewcommand{\arraystretch}{1.10}
\setlength{\tabcolsep}{6.0pt}
\begin{tabular}{@{}lllccc@{}}
\toprule
Schedule & Model & Router Variant & Full Accuracy (\%) & Harmful Episodes & $\Delta$ Accuracy \\
\midrule
Schedule~A & Qwen & \textbf{Full RC} & \textbf{70.2} & 9 & --- \\
 & & Without Global Advisor & 67.9 & 8 & --2.3 \\
 & & Without Local Advisor & 66.5 & 11 & --3.7 \\
 & & Without Meta Governor & 63.4 & 8 & --6.8 \\
\addlinespace[2.5pt]
 & GPT-4o & \textbf{Full RC} & \textbf{73.3} & 9 & --- \\
 & & Without Global Advisor & 71.3 & 6 & --2.0 \\
 & & Without Local Advisor & 67.3 & 19 & --6.0 \\
 & & Without Meta Governor & 66.8 & 6 & --6.5 \\
\midrule
Schedule~B & Qwen & \textbf{Full RC} & \textbf{71.6} & 8 & --- \\
 & & Without Global Advisor & 68.8 & 8 & --2.8 \\
 & & Without Local Advisor & 69.3 & 15 & --2.3 \\
 & & Without Meta Governor & 68.5 & 6 & --3.1 \\
\addlinespace[2.5pt]
 & GPT-4o & \textbf{Full RC} & \textbf{73.0} & 6 & --- \\
 & & Without Global Advisor & 70.5 & 6 & --2.6 \\
 & & Without Local Advisor & 70.7 & 14 & --2.3 \\
 & & Without Meta Governor & 70.2 & 5 & --2.8 \\
\bottomrule
\end{tabular}
\end{table}

\subsection{Metrics and sensitivity grids for \texorpdfstring{\(W_m\)}{Wm}, gate, and \texorpdfstring{\(W_g\)}{Wg}}
\label{app:metrics}
\emph{Why is the frozen configuration \((32,8,0)\) reasonable, and is the gain robust to it?}
Our primary diagnostic metric is accuracy on episodes where the Global
(\(W_g = 32\)) and Local (\(W_\ell = 1\)) proposal bits differ (\(g_t \neq
\ell_t\), identified without ground-truth labels), reported alongside its gain
over \(\max(G,L)\).
Key-cluster bootstrap resamples the 16 keys (5,000 replicates, seed 0) rather
than individual episodes, while point estimates are computed on the full stream.
On the disagreement slice of Table~\ref{tab:rc-main-dis}, bootstrap mean gains
span 4.3--6.2 points, with 95\% confidence intervals [0.0, 10.9] and [0.0, 9.6]
on Schedule~A (Qwen and GPT-4o) and [--0.6, 8.5] and [--0.8, 8.7] on Schedule~B
(positive in 93\%--97\% of replicates; joint resampling across all four runs
yields +5.1 [2.3, 8.1] on the disagreement slice and +1.9 [0.9, 3.1]
full-stream).
Tables~\ref{tab:mw-gate} and~\ref{tab:window-nbhd} report \emph{report-only}
sensitivity grids across \(W_m\), switching threshold, and \(W_g\); all
entries were computed after the frozen configuration was chosen and were never
used to tune \((32, 8, 0)\). The grids show that the gain is not knife-edge:
all \(W_m \in \{4, 8, 16\}\) and gate \(\in \{0, 1\}\) remain positive, and all
twelve \(W_g\) entries remain positive.

\begin{table}[h]
\centering
\caption{Report-only disagreement-slice gains (points) across arbitration window
size \(W_m\) and switching threshold on Schedule~A. Larger thresholds reduce the
gain; \((W_m = 8, \text{threshold } 0)\) is the frozen configuration.}
\label{tab:mw-gate}
\par\kern4pt
\footnotesize
\renewcommand{\arraystretch}{1.10}
\setlength{\tabcolsep}{6.5pt}
\begin{tabular}{@{}lcccccc@{}}
\toprule
 & \multicolumn{3}{c}{Qwen3-30B-A3B (Schedule~A)} & \multicolumn{3}{c}{GPT-4o (Schedule~A)} \\
\cmidrule(lr){2-4}\cmidrule(lr){5-7}
Window \(W_m\) & Gate 0 & Gate 1 & Gate 2 & Gate 0 & Gate 1 & Gate 2 \\
\midrule
4 & +8.2 & +6.4 & +1.8 & +6.0 & +3.7 & +3.0 \\
8 (frozen) & \textbf{+7.3} & +5.5 & +5.5 & \textbf{+5.2} & +4.5 & +3.7 \\
16 & +7.3 & +6.4 & +5.5 & +4.5 & +4.5 & +3.0 \\
\bottomrule
\end{tabular}
\end{table}

\begin{table}[h]
\centering
\caption{Report-only disagreement-slice gains (points) over \(\max(G,L)\) across
global window sizes \(W_g\). All twelve entries remain positive.}
\label{tab:window-nbhd}
\par\kern4pt
\footnotesize
\renewcommand{\arraystretch}{1.10}
\setlength{\tabcolsep}{8.0pt}
\begin{tabular}{@{}lcccc@{}}
\toprule
 & \multicolumn{2}{c}{Schedule~A} & \multicolumn{2}{c}{Schedule~B} \\
\cmidrule(lr){2-3}\cmidrule(lr){4-5}
Window \(W_g\) & Qwen & GPT-4o & Qwen & GPT-4o \\
\midrule
16 & +9.9 & +4.6 & +1.8 & +2.0 \\
32 (frozen) & \textbf{+7.3} & \textbf{+5.2} & \textbf{+5.3} & \textbf{+5.5} \\
64 & +5.7 & +4.9 & +3.1 & +4.4 \\
\bottomrule
\end{tabular}
\end{table}

\subsection{Scale and schedule-family robustness}
\label{app:scale}
\emph{Does the router still work at 54 keys and on schedule families absent from development?}
The audited LongMemEval knowledge-update pool yields 54 keys under the same
temporal-cue filter used for the main 16-key set.
To keep the per-key consolidation lag matched across pool sizes, we scale the
store refresh interval linearly with key count (\(6K\), or 96 episodes at
\(K = 16\) and 324 episodes at \(K = 54\)), giving each key six visits between
store updates (Table~\ref{tab:scale54}).
Table~\ref{tab:family} further evaluates two held-out schedule families absent
from development: Independent-Switch (one asynchronous flip per key at a hashed
step) and Poisson-Flip (repeated independent flips).
RegimeCouncil achieves positive gains across all eight runs; under Poisson-Flip,
opposing flips within \(\mathcal{W}_m\) keep \(A_t\) near zero so the Meta
Governor conservatively retains Local more often.

\begin{table}[h]
\centering
\caption{Disagreement-slice gain over \(\max(G,L)\) (points) on two held-out
schedule families (16-key pool, frozen windows).}
\label{tab:family}
\par\kern4pt
\footnotesize
\renewcommand{\arraystretch}{1.10}
\setlength{\tabcolsep}{7.5pt}
\begin{tabular}{@{}lcccc@{}}
\toprule
 & \multicolumn{2}{c}{Schedule~A} & \multicolumn{2}{c}{Schedule~B} \\
\cmidrule(lr){2-3}\cmidrule(lr){4-5}
Schedule Family & Qwen & GPT-4o & Qwen & GPT-4o \\
\midrule
Independent-Switch & \textbf{+6.7} & \textbf{+7.2} & \textbf{+6.9} & \textbf{+7.1} \\
Poisson-Flip & \textbf{+0.6} & \textbf{+3.1} & \textbf{+4.6} & \textbf{+3.3} \\
\bottomrule
\end{tabular}
\end{table}

\subsection{Governor window sweep and development comparisons}
\label{app:reactivity}
\emph{How sensitive is the Meta Governor to \(W_m\), and does it outperform symmetric trackers on development?}
Combining Schedules~A, B, and~C, the two held-out schedule families, the 54-key
pool, and both real-drift benchmarks across both LLMs yields 30 held-out
settings totaling 33,792 episodes.
Holding \((W_g, W_\ell, \text{gate}) = (32, 1, 0)\), Table~\ref{tab:reactivity}
sweeps \(W_m \in \{1, 2, 4, 8, 16, 32\}\) alongside Follow-recent-winner and
CUSUM on the same counterfactual surplus stream.
Every configuration outperforms \(\max(G,L)\) in all 30 of 30 settings, while
larger arbitration windows and symmetric trackers trade lower switch rates for
higher excess harmful exposure: \(W_m=8\) attains a median gap of +7.27 with
0.90 excess harm, whereas Follow-recent-winner attains +8.12 with 2.20 excess
harm. The frozen \(W_m=8\) therefore sits on the accuracy--harm frontier rather
than at the accuracy maximum.
Table~\ref{tab:dev-follow} reports the disjoint development comparison, where
the frozen \(W_m = 8\) rule achieves higher accuracy and fewer harmful episodes
than Follow-recent-winner across every development family.

\begin{table}[h]
\centering
\caption{Governor sensitivity across all 30 held-out settings (33,792
episodes). Gaps denote disagreement-slice gains over \(\max(G,L)\) in points
(upper-middle order statistic of the 30 sorted values; two-sided sample medians
appear in Table~\ref{tab:tie-meta-comp}). Switch Rate is the fraction of
consecutive disagreement steps on which the governor changes advisor; Excess
Harm is mean harmful episodes per setting beyond the safest fixed scope.}
\label{tab:reactivity}
\par\kern4pt
\footnotesize
\renewcommand{\arraystretch}{1.10}
\setlength{\tabcolsep}{6.0pt}
\begin{tabular}{@{}lccccc@{}}
\toprule
Governor Variant & Median Gap & Worst Gap & Settings $< 0$ & Switch Rate & Excess Harm \\
\midrule
\(W_m = 1\) & +3.97 & +0.62 & 0 / 30 & 0.093 & 0.13 \\
\(W_m = 2\) & +5.73 & +1.23 & 0 / 30 & 0.071 & 0.13 \\
\(W_m = 4\) & +7.09 & +1.85 & 0 / 30 & 0.049 & 0.27 \\
\textbf{\(W_m = 8\) (frozen)} & \textbf{+7.27} & \textbf{+0.62} & \textbf{0 / 30} & \textbf{0.035} & \textbf{0.90} \\
\(W_m = 16\) & +7.87 & +2.47 & 0 / 30 & 0.027 & 1.27 \\
\(W_m = 32\) & +7.20 & +1.85 & 0 / 30 & 0.024 & 2.33 \\
\addlinespace[2.5pt]
Follow-recent-winner & +8.12 & +2.47 & 0 / 30 & 0.025 & 2.20 \\
CUSUM on surplus & +7.94 & +1.23 & 0 / 30 & 0.019 & 2.20 \\
\bottomrule
\end{tabular}
\end{table}

\begin{table}[h]
\centering
\caption{Disjoint development comparison: disagreement-slice gain difference
(points) and total harmful episodes pooled over five development seeds and both
models (refresh interval 96).}
\label{tab:dev-follow}
\par\kern4pt
\footnotesize
\renewcommand{\arraystretch}{1.10}
\setlength{\tabcolsep}{7.0pt}
\begin{tabular}{@{}lccc@{}}
\toprule
Development Family & RC vs.\ Follow (points) & Harmful Episodes (RC) & Harmful Episodes (Follow) \\
\midrule
Default-Mixed & \textbf{+0.62} & \textbf{49} & 65 \\
Anticorrelated & \textbf{+0.47} & \textbf{38} & 46 \\
Long-Stable-Flip & \textbf{+0.07} & \textbf{94} & 105 \\
Sparse-Global-Transfer & \textbf{+0.53} & \textbf{17} & 28 \\
Block-Switch & \textbf{+0.84} & \textbf{69} & 79 \\
\bottomrule
\end{tabular}
\end{table}

\subsection{Naturally occurring drift details (TempLAMA and TAQA)}
\label{app:realdrift}
\emph{How are the real-drift streams constructed, and where does the large TempLAMA GPT-4o gap come from?}
TempLAMA \citep{dhingra2022time} extracts subject--relation--year facts from
Wikidata spanning 2010--2020; from the 2,191 single-valued fact groups that
change once, we sample a 120-key subset balanced across all nine relations.
TAQA \citep{zhao2024set} provides Wikipedia-table questions spanning 2000--2023;
from 1,000 validation items we retain the 284 items with a clean adjacent pair
of single-valued runs (each lasting at least two years) and hash-sample 100
keys.
Calendar years and explicit temporal cues are stripped so validity remains
unobservable at decision time (Assumption~\ref{ass:1}), and each key transitions
at the horizon fraction matching its true historical change year.
Tables~\ref{tab:realdrift} and~\ref{tab:taqa} report the individual per-schedule
results summarized in Table~\ref{tab:realmain}.

The +17.5 to +18.1 slice gap on TempLAMA with GPT-4o is larger than on the
synthetic streams because historical switch years cluster unevenly across
2010--2020: an early wave of switches turns Global's pooled window negative and
over-suppresses keys whose switch years arrive later, while one-step lag and
local ties hold back fixed Local during the early wave. The Meta Governor
overrides during the cluster and reverts to Local on still-valid keys, widening
the margin precisely where either fixed scope alone breaks down.

\begin{table}[h]
\centering
\caption{TempLAMA real-drift stream (120 keys, historical switch years, matched
refresh interval 720): disagreement-slice accuracy (\%). Full-stream gains are
+3.0 and +3.1 points on Qwen and +5.4 and +5.3 points on GPT-4o; module
ablations lower accuracy by 5.7--7.0 points without Local, 3.0--5.4 without
Global, and 3.9--5.8 without the Meta Governor.}
\label{tab:realdrift}
\par\kern4pt
\footnotesize
\renewcommand{\arraystretch}{1.10}
\setlength{\tabcolsep}{7.0pt}
\begin{tabular}{@{}llccccc@{}}
\toprule
Model & Schedule & $n_{\mathrm{dis}}$ & RC & Best G/L & Slice Gap & 95\% CI \\
\midrule
Qwen3-30B-A3B & Schedule~A & 1105 & \textbf{85.4} & 78.2 & \textbf{+7.2} & [5.8, 8.9] \\
 & Schedule~B & 1109 & \textbf{86.2} & 78.8 & \textbf{+7.4} & [6.0, 9.1] \\
\addlinespace[2.5pt]
GPT-4o & Schedule~A & 786 & \textbf{84.2} & 66.2 & \textbf{+18.1} & [15.2, 20.5] \\
 & Schedule~B & 793 & \textbf{83.6} & 66.1 & \textbf{+17.5} & [14.6, 19.8] \\
\bottomrule
\end{tabular}
\end{table}

\begin{table}[h]
\centering
\caption{TAQA real-drift stream (100 keys, historical switch years, matched
refresh interval 600): disagreement-slice accuracy (\%). Full-stream gains are
+3.4 and +3.5 points on Qwen and +3.7 and +3.8 points on GPT-4o; module
ablations lower accuracy by 3.9--5.3 points without Local, 3.4--3.8 without
Global, and 3.5--5.2 without the Meta Governor.}
\label{tab:taqa}
\par\kern4pt
\footnotesize
\renewcommand{\arraystretch}{1.10}
\setlength{\tabcolsep}{7.0pt}
\begin{tabular}{@{}llccccc@{}}
\toprule
Model & Schedule & $n_{\mathrm{dis}}$ & RC & Best G/L & Slice Gap & 95\% CI \\
\midrule
Qwen3-30B-A3B & Schedule~A & 856 & \textbf{83.4} & 74.6 & \textbf{+8.8} & [6.4, 10.8] \\
 & Schedule~B & 848 & \textbf{84.0} & 75.0 & \textbf{+9.0} & [7.0, 11.1] \\
\addlinespace[2.5pt]
GPT-4o & Schedule~A & 702 & \textbf{81.5} & 69.8 & \textbf{+11.7} & [9.1, 14.7] \\
 & Schedule~B & 682 & \textbf{82.4} & 70.1 & \textbf{+12.3} & [9.5, 15.6] \\
\bottomrule
\end{tabular}
\end{table}

\subsection{Accuracy--harm frontier}
\label{app:frontier}
\begin{figure}[h]
\centering
\IfFileExists{data/pareto_async.png}{%
  \includegraphics[width=0.98\linewidth]{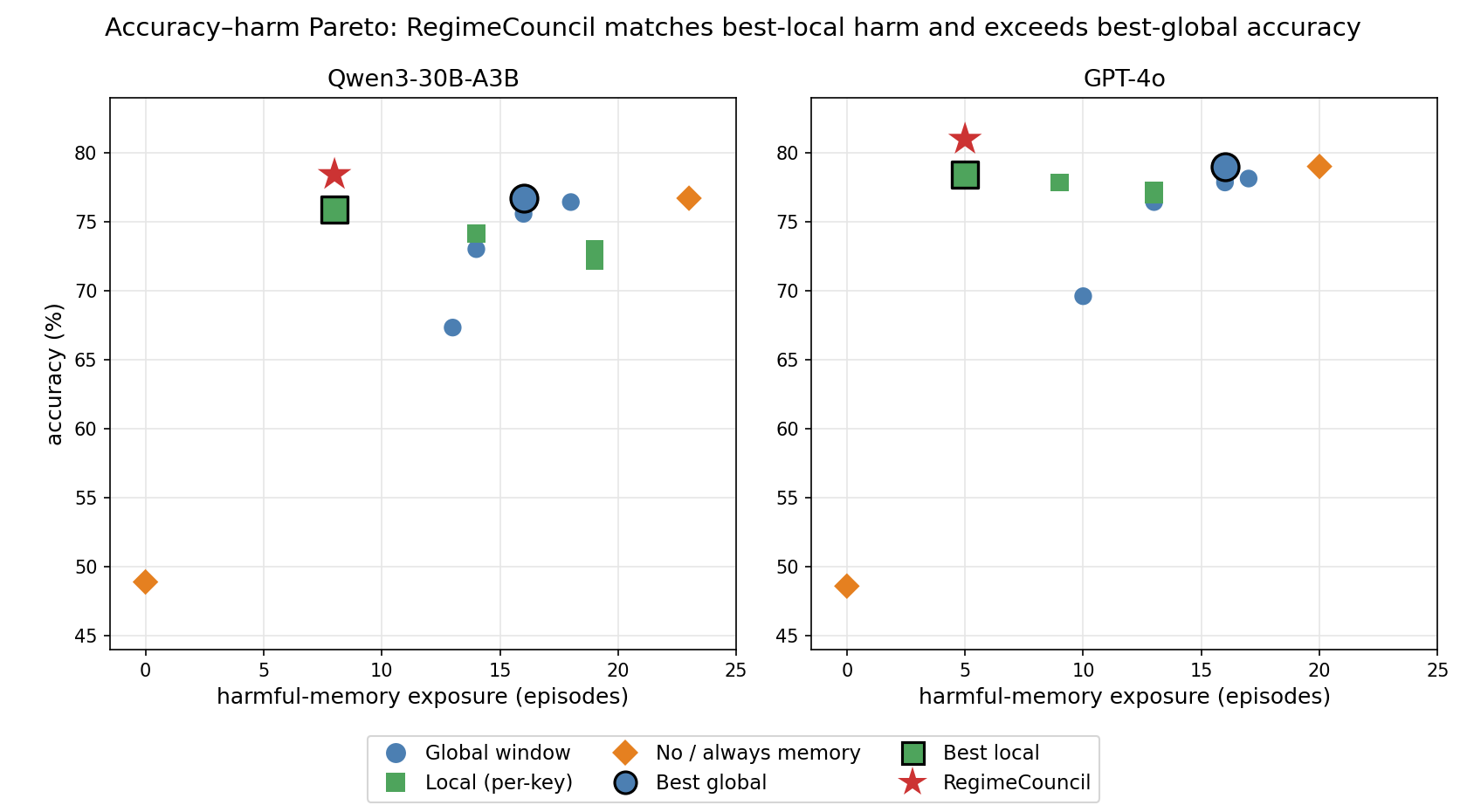}%
}{%
  \fbox{\parbox{0.9\linewidth}{\centering\vspace{6em}
  Pareto figure missing (\texttt{data/pareto\_async.png})\vspace{6em}}}%
}
\caption{Accuracy vs.\ harmful-memory exposure on the harm-exposure schedule.
Blue circles denote Global windows, green squares denote Local windows, orange
diamonds denote static no-memory and always-memory baselines, and the red star
marks RegimeCouncil; black outlines indicate the best Global and Local settings
in Table~\ref{tab:async-main}.}
\label{fig:pareto}
\end{figure}

\emph{Does the router trade accuracy for harm, or improve both?}
Table~\ref{tab:async-main} and Figure~\ref{fig:pareto} report the accuracy--harm
frontier on a dedicated harm-exposure schedule: RegimeCouncil matches the low
harmful exposure of the best Local window while surpassing the accuracy of the
best Global window on both models (+4.9 disagreement-slice points on Qwen with
\(n_{\mathrm{dis}} = 143\), and +6.4 points on GPT-4o with
\(n_{\mathrm{dis}} = 140\)).

\begin{table}[h]
\centering
\caption{Accuracy--harm trade-off on the dedicated harm-exposure schedule.}
\label{tab:async-main}
\par\kern4pt
\footnotesize
\renewcommand{\arraystretch}{1.10}
\setlength{\tabcolsep}{6.8pt}
\begin{tabular}{@{}llccc@{}}
\toprule
Model & Policy & Full Accuracy (\%) & Harmful Episodes & Max Drawdown \\
\midrule
Qwen3-30B-A3B & No memory & 48.9 & 0 & 0 \\
 & Always memory & 76.7 & 23 & 6 \\
 & Best Global (\(W_g = 16\)) & 76.7 & 16 & 2 \\
 & Best Local (\(W_\ell = 1\)) & 75.9 & 8 & 1 \\
 & \textbf{RC} & \textbf{78.4} & \textbf{8} & \textbf{1} \\
\addlinespace[2.5pt]
GPT-4o & No memory & 48.6 & 0 & 0 \\
 & Always memory & 79.0 & 20 & 2 \\
 & Best Global (\(W_g = 16\)) & 79.0 & 16 & 2 \\
 & Best Local (\(W_\ell = 1\)) & 78.4 & 5 & 1 \\
 & \textbf{RC} & \textbf{81.0} & \textbf{5} & \textbf{1} \\
\bottomrule
\end{tabular}
\end{table}

\subsection{Observable-conflict selection (supplementary baseline check)}
\label{app:mab}
\emph{Supplementary check: does the arbitration rule still help when validity is observable rather than hidden?}
This subsection is a supplementary baseline check outside our main
hidden-validity setting and is not used to support the main claims.
FactConsolidation in MemoryAgentBench \citep{hu2026memoryagentbench} places
conflicting writes alongside explicit ordering cues in the prompt, making
validity observable rather than hidden.
Table~\ref{tab:mab-main} evaluates certified rule selection (Bonferroni
\(0.05/3\) on 132 certification items, tested on 68 held-out items) and online
surplus arbitration (\(W_m = 8\), gate 0) between generic retrieval and
last-write-wins (LWW).

\begin{table}[h]
\centering
\caption{Observable-conflict rule selection on MemoryAgentBench 64K (held-out
accuracy \%, \(n = 68\)). Gains over Gen@5 are +35.3 points on GPT-4o and +25.0
points on Qwen.}
\label{tab:mab-main}
\par\kern4pt
\footnotesize
\renewcommand{\arraystretch}{1.10}
\setlength{\tabcolsep}{5.0pt}
\begin{tabular}{@{}lccccccc@{}}
\toprule
 & \multicolumn{4}{c}{Candidate Retrieval Rules} & \multicolumn{3}{c}{Selection \& Arbitration} \\
\cmidrule(lr){2-5}\cmidrule(lr){6-8}
Model & Gen@5 & Gen@20 & LWW@5 & LWW@20 & Selected Rule & Certified & Online Meta \\
\midrule
GPT-4o & 19.1 & 20.6 & 55.9 & 54.4 & LWW@20 & \textbf{54.4} & \textbf{54.4} \\
Qwen3-30B-A3B & 17.6 & 17.6 & 42.6 & 39.7 & LWW@5 & \textbf{42.6} & 41.2 \\
\bottomrule
\end{tabular}
\end{table}

\subsection{Artifact index}
\label{app:artifacts}
\emph{What will be released?}
Code and evaluation artifacts will be publicly released, including the
stream builders (the 16-key and 54-key LongMemEval streams and the TempLAMA and
TAQA real-drift benchmarks), cached model completions, the frozen router
implementation (conservative RC and RC-TD across paired
and bandit modes), and scripts reproducing every table in the paper. The
SHA-256 item-selection seed is included so the confirmatory set can be
regenerated independently of any router evaluation.

\section{Prompt templates}
\label{app:prompts}

RegimeCouncil does not prompt the LLM to choose between Global and Local scopes:
the model only answers the user query, while the router acts deterministically
on anchor-referenced utilities.
We therefore cache three completions per key and hold the prompt fixed across
all routing policies.
For the main 16-key and 54-key caches, the optional snapshot line is omitted
for the no-memory completion and populated with either the old or updated write
for the two memory completions; display order of the two conflicting records is
arbitrary and carries no recency cue.

\begin{promptbox}
\raggedright
\texttt{Two memory records conflict. Their display order is arbitrary and
the currently valid regime is hidden. Return only the short answer.}\\[0.4em]
\texttt{\{collision\_observation\}}\\[0.4em]
\texttt{A deployed feedback snapshot records the correct answer for this
key as: \{snapshot\}}\\[0.15em]
\textit{[snapshot line included only for memory completions; omitted for no-memory]}\\[0.4em]
\texttt{Question: \{question\}}
\end{promptbox}

A concrete instantiation for the volleyball-record item (abridged collision
text) is shown below:

\begin{promptbox}
\raggedright
\texttt{Two memory records conflict. Their display order is arbitrary and
the currently valid regime is hidden. Return only the short answer.}\\[0.4em]
\texttt{Conflicting memory A: [\ldots] I've been doing pretty well in the
volleyball league, we're 3-2 so far!}\\[0.3em]
\texttt{Conflicting memory B: [\ldots] our volleyball team, the Net Ninjas,
is doing well with a 5-2 record.}\\[0.4em]
\texttt{Question: What is my current record in the recreational volleyball
league?}
\end{promptbox}

Without a snapshot, the model must guess from two unordered records; with a
snapshot, it receives one of 3-2 or 5-2 as the latest committed write for that
key, which may already be stale relative to the hidden valid state.
Because the identical prompt template and cached completions are shared by the
Global Advisor, the Local Advisor, and RegimeCouncil, prompt wording cannot
favor one routing policy over another, and standard RAG pipelines exhibit the
same staleness vulnerability under neutral retrieval headers on LongMemEval
\citep{wu2024longmemeval}.
On the supplementary MemoryAgentBench check (Appendix~\ref{app:mab}), generic
top-\(k\) retrieval uses ``Answer from the retrieved facts,'' whereas
last-write-wins additionally states that larger serial numbers indicate newer
writes.

\section{Worked disagreement episodes}
\label{app:cases}

Table~\ref{tab:cases} traces two Schedule~A episodes on Qwen3-30B-A3B where the
Global and Local Advisors disagree.
Utilities equal 1 if the deployed completion matches the hidden ground truth and
0 otherwise; ``Use memory'' deploys the committed snapshot and ``Skip memory''
falls back to the no-memory anchor.

\begin{table}[h]
\centering
\caption{Two illustrative advisor-disagreement episodes on Schedule~A
(Qwen3-30B-A3B). The Global Advisor pools the 32 most recently revealed paired
differences across keys, the Local Advisor reads only the queried key, and the
Meta Governor overrides Local iff \(A_t > 0\).}
\label{tab:cases}
\par\kern4pt
\footnotesize
\renewcommand{\arraystretch}{1.12}
\setlength{\tabcolsep}{4.8pt}
\begin{tabular}{@{}lcccc@{}}
\toprule
Episode (Valid Answer) & Global & Local & Meta Decision & RC \\
\midrule
Volleyball record (3-2) & Use memory & Skip memory & Override (\(A_t > 0\)) & \textbf{Use memory (correct)} \\
Spare screwdriver (No) & Skip memory & Use memory & Retain Local (\(A_t \le 0\)) & \textbf{Use memory (correct)} \\
\bottomrule
\end{tabular}
\end{table}

At visit 7 of the volleyball-record item (where the committed write 3-2 remains
valid), using memory scores 1 whereas skipping scores 0: conservative Local
abstains after an earlier tie, while Global has accumulated positive evidence
from other keys and \(A_t > 0\) triggers a helpful override.
Conversely, at visit 14 of the spare-screwdriver item (where the committed
write ``No'' remains valid), Local's latest outcome on this key is positive
whereas mixed flips on other keys turn the pooled window non-positive; because
\(A_t \le 0\), the Meta Governor retains Local and answers correctly, whereas
Global alone would have abstained.

\section{Extended delay, consolidation-lag, and drift-shift analyses}
\label{app:extended-analysis}

This appendix presents the full store-refresh sweep, mid-stream drift-rate
shift evaluations, and executed-action bandit and runtime profiling details
supporting Sections~\ref{sec:module-ablation}--\ref{sec:delay-bandit}, all using
the same frozen \((W_g, W_\ell, W_m, \text{gate}) = (32, 1, 8, 0)\)
configuration.

\subsection{Store refresh interval vs.\ outcome revelation delay}
\label{app:delay-commit}
\phantomsection\label{app:tie-deferral}%
\phantomsection\label{tab:tau-sweep}%
Whereas Panel~A of Table~\ref{tab:delay-bandit-main} varies the outcome
revelation delay \(\tau \in [1, 96]\) at a fixed store refresh interval of
\(6K\), Table~\ref{tab:commit-sweep} varies the refresh interval at fixed
\(\tau = 1\) across Schedules~A, B, and~C on both models.
Because each of the 16 keys is visited once per 16-episode pass, refresh
intervals of 16 episodes or less write updates back before a key's next visit;
in that regime, stored snapshots are already current by the next query and
always using memory attains 93.0\% accuracy.
Across multi-pass consolidation lags of 32 to 192 episodes (2 to 12 passes
between store refreshes), stale snapshots persist across multiple queries:
conservative RegimeCouncil improves over always using memory by +0.5 to +5.5
full-stream points from 4 passes onward, and RC-TD improves across
every multi-pass interval (+0.1 to +5.6 points).

Under multi-step revelation delays \(\tau > 1\) (Table~\ref{tab:delay-bandit-main},
Panel~A), Proposition~\ref{prop:A}(ii) predicts that per-key regret scales with
\(\lceil \tau / K \rceil\), the maximum number of visits to a key before its
post-switch outcome arrives: at \(\tau = 16\), each key experiences one delayed
visit on the 16-key pool (\(\tau / K = 1\)), compared with at most one delayed
visit on fewer than a third of keys on the 54-key pool (\(\tau / K = 0.30\))
and \(\tau / K \le 0.16\) on TempLAMA and TAQA.
Once \(\tau \ge K\), many keys under conservative RegimeCouncil remain undecided
(\(\ell_t = 0\)) for extended stretches; deferring undecided local keys to
\(g_t\) in RC-TD restores positive gains across all 126 of 126
pool--delay settings up to \(\tau = 96\).

\begin{table}[ht]
\centering
\caption{Varying the store refresh interval at fixed outcome delay \(\tau = 1\)
across Schedules~A, B, and~C on both models (mean full-stream accuracy, \%).
Refresh intervals of 16 episodes or less correspond to single-pass write-back
before a key's next visit.}
\label{tab:commit-sweep}
\par\kern4pt
\footnotesize
\renewcommand{\arraystretch}{1.12}
\setlength{\tabcolsep}{3.0pt}
\begin{tabular}{@{}lcccccc@{}}
\toprule
 & \multicolumn{4}{c}{Mean Full-Stream Accuracy (\%)} & \multicolumn{2}{c}{Gain vs.\ Always Memory} \\
\cmidrule(lr){2-5}\cmidrule(lr){6-7}
Store Refresh Interval & Always Memory & Best G/L & RC & \textbf{RC-TD} & RC & \textbf{RC-TD} \\
\midrule
1 or 16 episodes (1 pass, write-back) & \textbf{93.0} & 92.6 & 91.6 & 92.7 & --1.4 & --0.3 \\
32 episodes (2 passes) & 87.7 & 86.4 & 86.9 & \textbf{87.8} & --0.8 & \textbf{+0.1} \\
64 episodes (4 passes) & 78.9 & 77.5 & \textbf{79.4} & \textbf{80.3} & \textbf{+0.5} & \textbf{+1.4} \\
96 episodes (6 passes, primary) & 70.1 & 69.6 & \textbf{71.9} & \textbf{72.8} & \textbf{+1.8} & \textbf{+2.7} \\
192 episodes (12 passes) & 51.2 & 56.7 & \textbf{56.7} & \textbf{56.8} & \textbf{+5.5} & \textbf{+5.6} \\
\bottomrule
\end{tabular}
\end{table}

\subsection{Mid-conversation drift-rate shifts, bandit details, and runtime profiling}
\label{app:feedback-cost}
\phantomsection\label{tab:realistic-cost}%
Beyond stationary drift rates, user attributes can alternate between stable
stretches and rapid bursts of change within a single interaction stream.
Table~\ref{tab:drift-shapes} evaluates frozen RegimeCouncil when the per-visit
flip hazard \(p_t\) shifts mid-stream across the 16-key and 54-key pools and
both LLMs: RegimeCouncil outperforms \(\max(G,L)\) across every tested hazard
range and trajectory shape, including two-stage step changes in both directions,
three-stage escalations, and continuous sinusoidal hazard waves.

\begin{table}[ht]
\centering
\caption{Robustness under mid-conversation drift-rate variation with frozen
windows across the 16-key and 54-key pools and both models: mean full-stream
accuracy (\%) and median disagreement-slice gain over \(\max(G,L)\) (points).}
\label{tab:drift-shapes}
\par\kern4pt
\footnotesize
\renewcommand{\arraystretch}{1.12}
\setlength{\tabcolsep}{3.0pt}
\begin{tabular}{@{}lccc@{}}
\toprule
In-Conversation Drift-Rate Trajectory (\(p_t\)) & Full Accuracy (\%) & Median Slice Gap & Settings $> 0$ \\
\midrule
Two-stage shift between 0.04 and 0.30 (both directions) & \textbf{68.1} & \textbf{+4.3} & 16 / 16 \\
Moderate two-stage shift between 0.02 and 0.20 & \textbf{70.5} & \textbf{+4.4} & 16 / 16 \\
High-volatility shift between 0.05 and 0.40 & \textbf{63.7} & \textbf{+2.5} & 15 / 16 \\
Three-stage escalation (\(0.02 \to 0.14 \to 0.35\)) & \textbf{66.2} & \textbf{+2.4} & 8 / 8 \\
Continuous sinusoidal wave (\(p_t \in [0.02, 0.35]\)) & \textbf{67.2} & \textbf{+2.7} & 8 / 8 \\
\bottomrule
\end{tabular}
\end{table}

In the executed-action bandit evaluation of Table~\ref{tab:delay-bandit-main}
(Panel~B), at each step \(t\) the router computes its deterministic proposal
\(\tilde{a}_t \in \{0, 1\}\) and deploys \(a_t = \tilde{a}_t\) with probability
\(1 - \varepsilon\) and \(1 - \tilde{a}_t\) with probability \(\varepsilon\)
(\(\varepsilon = 0.05\), averaged over 5 exploration seeds across all 30
held-out settings, or 150 runs in total).
Only the single deployed action \(a_t\) is executed and scored (\(u_t \in \{0,
1\}\)), with no background anchor call.
Feeding the inverse-propensity estimates \(\hat{d}_s\) and
\((\hat{u}_{s,\mathrm{mem}}, \hat{u}_{s,0})\) from Section~\ref{sec:paired} into
the unchanged \((32, 1, 8, 0)\) windows yields 68.9\% mean full-stream accuracy
and 31.3 harmful episodes per stream for RegimeCouncil, outperforming Global
(63.2\% accuracy, 60.6 harmful episodes) in all 30 settings,
Follow-recent-winner (68.5\%, prevailing in 23 of 30 settings, two-sided sign-test
\(p = 0.002\)), and Local (68.5\%, prevailing in 18 of 30 settings).
Single-threaded Python profiling over all 33,792 held-out decisions records
0.88\,\textmu s per decision for the Global Advisor alone and 2.29\,\textmu s
per decision for full RegimeCouncil (+1.41\,\textmu s overhead, or 440,000
decisions/s), adding zero user-facing critical-path latency.

\end{document}